\documentclass[Afour,sageh,times]{sagej}

\usepackage{moreverb,url}
\usepackage{graphicx, float}
\usepackage[ruled,vlined,linesnumbered]{algorithm2e}
\usepackage{amsmath}
\usepackage{color,soul}
\usepackage{xcolor}
\usepackage{subcaption}
\usepackage{flushend}
\usepackage{adjustbox}
\usepackage{tabularx}
\usepackage{enumitem}

\newcounter{myenumi}

\renewcommand\theenumi{\arabic{myenumi}. }

\long\def\Item#1\par{%
 \stepcounter{myenumi}%
 \makebox[1.5em]{\hl\theenumi}\hl{ #1}%
 \par
}

\renewcommand{\themyenumi}{
\setlength{\parindent}{0pt}
\arabic{myenumi}.}

\newenvironment{myenumerate}{%
\setlength{\parindent}{0pt}
\setcounter{myenumi}{0}
\renewcommand{\item}{
\par
\refstepcounter{myenumi}
\makebox[1.5em][l]{\themyenumi}
}
}{
\par
\noindent
\ignorespacesafterend
}
\usepackage[colorlinks,bookmarksopen,bookmarksnumbered,citecolor=red,urlcolor=red]{hyperref}

\newcommand\BibTeX{{\rmfamily B\kern-.05em \textsc{i\kern-.025em b}\kern-.08em
T\kern-.1667em\lower.7ex\hbox{E}\kern-.125emX}}

\def\volumeyear{2019}

\begin{document}

\runninghead{Sabbagh Novin et al.}

\title{A Model Predictive Approach for Online Mobile Manipulation of Nonholonomic Objects using Learned Dynamics}

\author{Roya Sabbagh Novin\affilnum{1} Amir Yazdani\affilnum{1} Andrew Merryweather\affilnum{1} Tucker Hermans\affilnum{2} }

\affiliation{\affilnum{1}Department of Mechanical Engineering and Utah Robotics Center,
        University of Utah, Utah, US\\
\affilnum{2}School of Computing and Utah Robotics Center, University of Utah, Utah, USA}

\corrauth{Roya Sabbagh Novin, Department of Mechanical Engineering,
        University of Utah, Utah, US.}

\email{roya.sabbaghnovin@utah.edu}

\begin{abstract}
Assistive robots designed for physical interaction with objects will play an important role in assisting with mobility and fall prevention in healthcare facilities. Autonomous mobile manipulation presents a hurdle prior to safely using robots in real-life applications. In this article, we introduce a mobile manipulation framework based on model predictive control using learned dynamics models of objects. We focus on the specific problem of manipulating legged objects such as those commonly found in healthcare environments and personal dwellings (e.g. walkers, tables, chairs). We describe a probabilistic method for autonomous learning of an approximate dynamics model for these objects. In this method, we learn dynamic parameters using a small dataset consisting of force and motion data from interactions between the robot and object. Moreover, we account for multiple manipulation strategies by formulating manipulation planning as a mixed-integer convex optimization. The proposed framework considers the hybrid control system comprised of i) choosing which leg to grasp, and ii) control of continuous applied forces for manipulation. We formalize our algorithm based on model predictive control to compensate for modeling errors and find an optimal path to manipulate the object from one configuration to another. We show results for several objects with various wheel configurations. Simulation and physical experiments show that the obtained dynamics models are sufficiently accurate for safe and collision-free manipulation. When combined with the proposed manipulation planning algorithm, the robot successfully moves the object to the desired pose while avoiding any collision.
\end{abstract}


\maketitle

\section{Introduction}
The lack of reliable and safe mobile manipulation algorithms prevents robots from performing complicated manipulations in unstructured environments (\cite{stilman2007manipulation, dogar2011framework}) or, more importantly, in close proximity to humans, such as in healthcare environments or warehouses. The ultimate goal of autonomous mobile manipulation is to perform complex manipulation tasks in dynamic environments. The manipulation task is defined as moving an object from an initial configuration to a given goal configuration~(\cite{berenson2008optimization, ciocarlie2012mobile}). The reliability of the manipulation planning significantly decreases in the case of manipulating objects with unknown dynamics models. In this paper, we focus on the problem of online mobile manipulation of legged objects using learned dynamics models of objects and present an optimization-based framework for mobile manipulation planning.

\begin{figure}[t!]
\center
\includegraphics[width = 8.5cm]{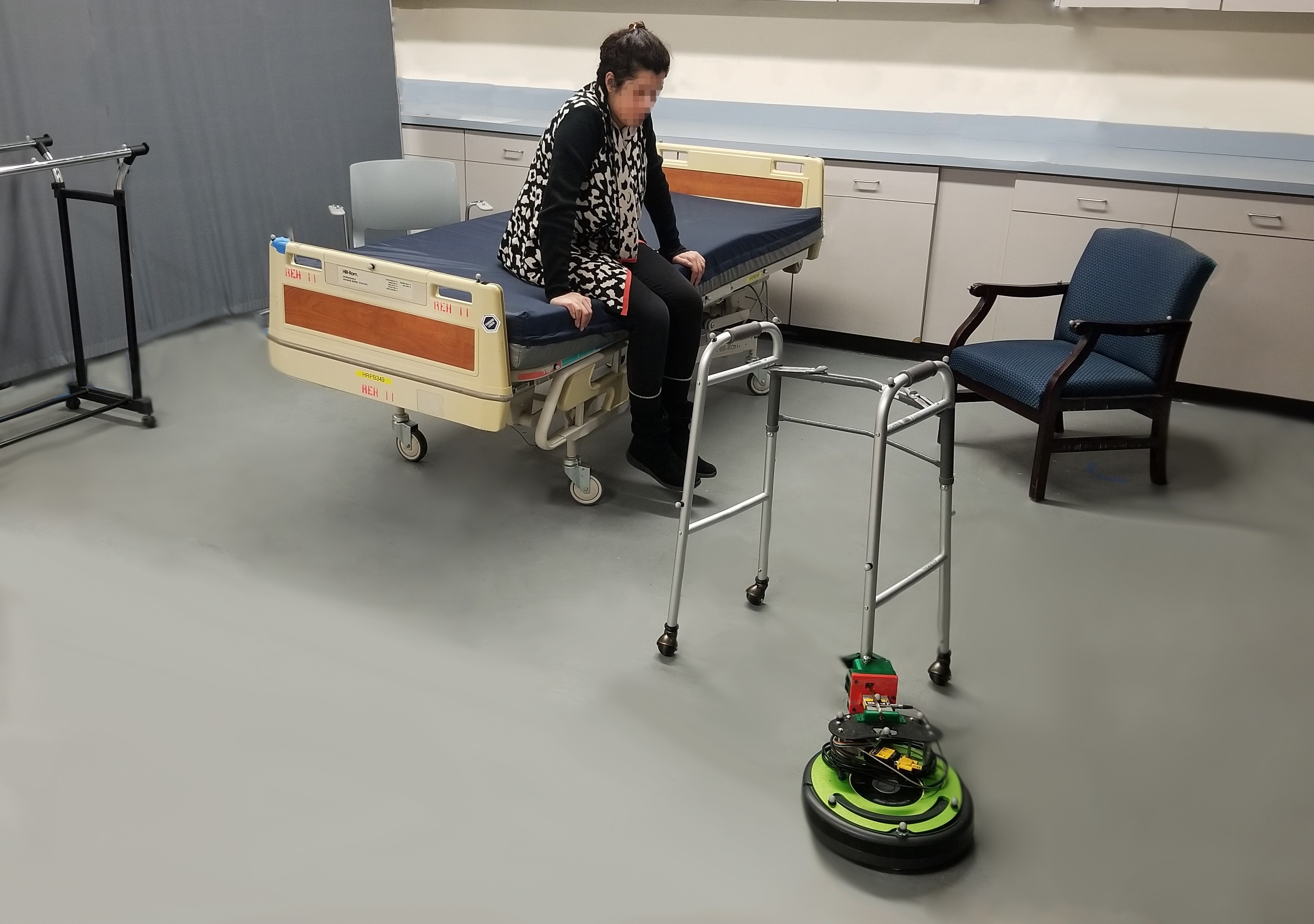}
\caption{Hospital room setup for experiments. The planning algorithm considers both pushing and pulling objects which is helpful in small spaces and corners, when attempting to deliver or retrieve the walking aid.}
\label{Real robot}
\vspace{-0.3in}
\end{figure}

One significant potential application of mobile manipulation planning is in healthcare environments. The robot's ability to manipulate objects in healthcare environments would significantly increase the number of tasks in which robots can play a role, thus freeing professional care providers to focus on tasks requiring their unique expertise. For instance, falls with serious injury are consistently among the top 10 sentinel events with the majority of these falls occurring in hospitals~(\mbox{\cite{Chu2017, Joint2015}}). One of the most significant healthcare factors related to falls is the nurse-to-patient ratio~(\cite{Chu2017}). In the case of short staffing/nurse workload, even when an alarm activates, it may take minutes for a nurse to respond, and falls often happen during this response time when patients have little to no support (\cite{kristoffersson2013review, oliver1997development}). 

Patient sitters are one of the solutions to overcome this problem. However, a patient sitter could be replaced with assistive robots. We believe an autonomous assistant mobile robot with object manipulation capabilities can prevent patient falls by intervening with a mobility aid at the bedside. The robot uses monitoring data to plan assistance by providing a mobility aid to a patient or clears the patient's path by moving obstacles away (Fig.~\ref{Real robot}).

Safe autonomous mobile manipulation is also beneficial in other applications such as warehousing and intralogistics~\mbox{(\cite{fragapane2020increasing})}. It reduces costs and improves efficiency and productivity while working alongside people \mbox{(\cite{roa2015mobile})}.

One of the main challenges in using robots for such applications is safety. A robot maneuvering in close proximity to humans needs to consider human motion and intention to avoid any collision. The problem of human-aware autonomous mobile robot navigation in unstructured and dynamic environments has been widely investigated~(\cite{nakhaeinia2015hybrid, kretzschmar2016socially, pol2015review}). However, many challenges still exist in manipulating objects while navigating through unstructured environments such as hospitals or personal dwellings~(\cite{nakhaeinia2015hybrid, kretzschmar2016socially, pol2015review, mast2015design}). 

\begin{figure}[t!]
\center
\includegraphics[width = 8.5cm]{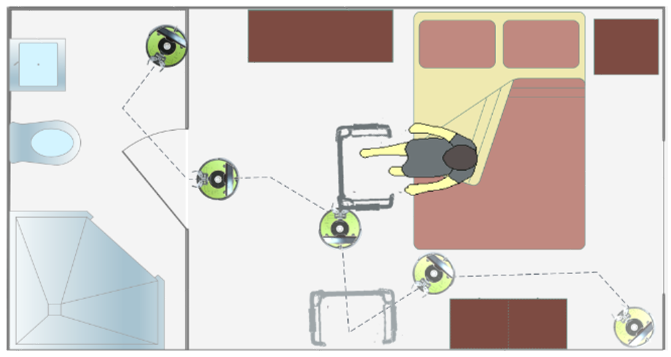}
\caption{Whenever needed, the robot can retrieve the walker and bring it to the patient to help with stability to prevent a fall.}
\label{Room}
\vspace{-0.25in}
\end{figure}

The main challenges include estimating unknown objects' dynamics, creating safe and collision-free maneuvering trajectories, and dealing with discrete and continuous, i.e., hybrid actions. Medical environments are usually cluttered by various objects, including mobility aids, carts, chairs, and tables. Since these objects' dynamics are not necessarily known to the robot, the robot must be able to determine the object's dynamics properties autonomously for successful manipulation. Additionally, when manipulating legged objects, the robot must select not only a direction and magnitude of the pushing or pulling force, but also make a discrete choice of which leg to push or pull. Thus, in this paper, we investigate the problem of manipulating unknown legged objects to a desired final position using our customized robot (Fig. \ref{Room}). 

Previous research concerning dynamics models mostly describe either mapping between actions and consequences for a specific task (\cite{ogata2005extracting,fitzpatrick2003learning}), or rely on pure kinematics~(\cite{vithani2002estimation}). A more advanced approach mimics human sensorimotor learning behavior, in which a coarse dynamics model of the new object is learned based upon prior beliefs and experiences. The coarse model is eventually improved as more data are collected during the manipulation~(\cite{scholz2015learning,kording2004bayesian}). In this research, we choose a Bayesian regression model to incorporate knowledge about common legged furniture as priors to inform the dynamics learning algorithm~(\cite{kording2004bayesian}).

The main contribution of this paper is the hybrid manipulation planning in which we provide a framework to manipulate large legged objects with complicated dynamics. We formulate the problem as a mixed-integer convex optimization to solve the hybrid control problem comprised of i) choosing which leg to grasp, and ii) continuous applied forces to move the object. 

We learn simplified dynamics models and incorporate them into the online manipulation planning where there is no prior knowledge of action sequences as used in the state-of-the-art literature (\mbox{\cite{hogan2017reactive, pajarinen2017hybrid}}). The reason that we use a learning method to fit the models is because pure analytic models of complex dynamics are difficult to obtain for accurate planning. We implemented three different models to investigate the effects of (1) including the intertia in the model, and (2) adding a second wheel and using a more complicated model. We used these models to show how much difference they make in the prediction performance. We use model predictive controller (MPC) to overcome the imperfections of the dynamics model and avoid getting stuck in local minima. To make it convex, we linearize the dynamics model over a nominal trajectory for the object.

We use a gripper instead of just pushing objects with the robot. The main reason for this is because the potential points of contact in legged objects are limited, and by just pushing, we could easily loose contact. Thus, We used a gripper to partially immobilize the object relative to the robot and keep it in contact. In addition, Using a gripper gives the robot the ability to also pull whenever it is easier, faster or pushing is not feasible. Our designed gripper acts as an instantaneous revolute joint, meaning that it does not exert any torque to the object and the object's leg can rotate inside the gripper.

We divide the mobile manipulation problem into four sub-problems: 1) move the robot from its initial state to a state where it is near the object and can move to a grasping state, 2) move to a grasping state and grasp, 3) move the robot (grasping the object) in such a way that it moves the object into its goal configuration, 4) re-position the robot whenever it gets close to performance limits.
 
Below, we summarize the main contributions of this paper:
\begin{myenumerate}
\item Unknown Object Dynamics Learning: Using the Bayesian regression method adopted from \cite{scholz2016navigation} to learn dynamic parameters of legged objects and investigate three different dynamics models using experimental data. 
\item Hybrid Manipulation Planner: Developing a manipulation planner based on receding horizon concept and mixed-integer convex optimization with discrete actions of changing legs as well as continuous motion in manipulation.
\end{myenumerate}

This paper builds on the preliminary results presented in the conference paper (\cite{novin2018dynamic}), where the robot design, the procedure for object parameter estimation, and the main concepts of manipulation planning algorithm were introduced. The major improvements with respect to this previous work are:
\begin{myenumerate}
\item We introduce an additional mode for grasping, which reduces multiple efforts to grasp a leg and makes overall object manipulation faster.

\item We develop a repositioning mode, which prevents the robot from getting stuck in between two legs of the object.

\item We investigate more complicated dynamics models and run the dynamics model learning on several objects.

\item We use a new system of weight assignment for the optimization cost function.

\item We perform simulation and physical experiments using the proposed manipulation planning algorithm on multiple legged objects.

\end{myenumerate}

We organize the remainder of the paper as follows. We discuss an overview of related work in Section 2. In Section 3, we introduce and formalize the mobile manipulation problem. We follow this with details of our approach in Section 4, 5 and 6, including the robot model, dynamics parameter learning method and manipulation planning algorithm. Section 7 explains the implementation aspects of Bayesian regression on collected data using a real robot and the experimental protocol for evaluating our object manipulation algorithm. We analyze the results of extensive robot experiments in Section 8. Finally, a closing discussion and potential future work are presented in Section 9.

\section{Related Work}

\subsection{Assistive robots}
Most assistive robots developed to support independent living are only used to monitor, communicate, or deliver supplies in hospitals without any physical engagement with patients (e.g., Giraff by \cite{pripfl2016results} and HOBBIT by \cite{casiddu2015robot}). However, these robots have the potential to do more interactive tasks that are repetitive, time-consuming, and do not need the expertise of a professional care provider. \cite{Chen2013} have developed assistive capabilities for the PR2 robot to empower people with severe motor impairments to interact with the physical world. They have investigated various tasks through two case studies, including scratching and shaving, retrieving an object at home, and socially interacting through speech and gesture. In all these tasks, the robot is directly controlled by the human.

We believe that assistive robots would significantly benefit from an autonomous manipulation planning framework, which is the subject of this paper. By enabling robots to perform manipulation tasks, we can reduce the workload of healthcare providers and improve the quality of senior independent living.

\subsection{Object model identification in motion planning}
The estimation of dynamic parameters of a manipulated object by an autonomous mobile robot has received some attention in the past \cite{stuber2020let}. Most of the existing approaches either require an extensive training dataset~(\cite{fitzpatrick2003learning}) or use kinematics-based methods for a specific task~(\cite{vithani2002estimation}).

Some studies are based on learning a mapping between actions and the resulting motion to describe an object's dynamic behavior and inform future goal-directed behavior (\cite{ogata2005extracting,fitzpatrick2003learning, zhou2018convex}). \mbox{\cite{li2018push}} uses a deep recurrent neural network model to learn object motion properties for planar pushing for a single object. \mbox{\cite{yuan2018rearrangement}} design a learning system that treats perception, action planning, and motion planning in an end-to-end process. \mbox{\cite{mericcli2015push}} learns case-based planar motion of objects on the plane which is object-specific. The most limiting drawback in these data-driven methods is that since they do not provide any physics-based dynamics model, they cannot be used for other manipulation tasks other than what is performed in the training process. Thus, they are very limited in terms of handling new task requirements and usually suffer from high uncertainty in the real-world experiments.

Other studies suggest finding dynamic parameters of objects to overcome the limitations of mapping methods. \cite{stilman2007learning} used the pseudo-inverse of dynamics equations to obtain the dynamic parameters of large objects modeled as ``a point mass on a wheel". However, they were not successful in finding a consistent relationship between acceleration and force and only used a viscous friction model, ignoring mass and inertia parameters.

Later, learning methods were used to estimate non-linear dynamics models of objects. \mbox{\cite{bauza2018data}} used a data-driven approach to model planar pushing interaction for model-predictive control, but their learned model is a task-specific Gaussian Process. \cite{scholz2015learning} used physics-based reinforcement learning as an adaptive method to obtain dynamics models of nonholonomic objects. They used this method to estimate the physical parameters of an office table and a utility cart with fixed front wheels using the same ``point mass on a wheel" model~(\cite{scholz2016navigation}).

Least squares approaches are also used for object kinematic and dynamic parameter estimation~(\cite{Cehajic2017}). Others use interaction data between a team of mobile robots and object to find mass and inertia parameters~(\cite{Marino2018, Franchi2015}). More complicated models are also investigated for nonholonomic objects. \cite{sun2002interactive} use least squares to identify model parameters of a 4-wheel cart manipulated by a mobile manipulator. However, they ignore the friction effects. 

In this paper, we adopt the method proposed by \mbox{\cite{scholz2015learning}}, which is a combination of learning and analytical modeling to obtain dynamic parameters of various real objects. We run real-world experiments and obtain dynamics parameters using three different models and choose the one that can predict most accurately.

\subsection{Manipulation planning}
There is also an ongoing effort to find planning frameworks that can effectively handle the uncertainty and hybrid properties associated with planning for pushing and pulling actions (\cite{lynch1996stable}). \cite{mason1986mechanics} first formulated the mechanics of planar pushing manipulation tasks. \cite{salganicoff1993vision} created a forward empirical model of an unknown object for pushing using visual feedback. \cite{li2007manipulation} focused on finding appropriate pushing actions and developing a push planner that can track a predefined trajectory using these actions based on a set of assumptions and a simplified model of two-agent point-contact push. \mbox{\cite{zito2012two}}
proposed a two-stage approach, including a global RRT path planner and a local push planner. They use predictive models of pushing interactions to plan sequences of pushes to move an object in an obstacle-free tabletop scenario. The local planner utilized a physics engine that requires explicit object modeling.

\cite{arruda2017uncertainty} used a model predictive path integral controller to plan push manipulations based on a learned model including uncertainties, obtained by Gaussian process regression and an ensemble of mixture density networks. \cite{hermans2013learning} presented a data-driven approach for learning good contact locations for pushing unknown objects. \mbox{\cite{krivic2019pushing}} suggested a multi-layer, modular method for pushing unknown objects in cluttered and dynamic environments. They introduce pushing corridors and use an adaptive pushing controller which learns local inverse models of robot-object interaction online.

\cite{desai1997nonholonomic} addressed the problem of motion planning for nonholonomic cooperating mobile robots manipulating and transporting objects while holding them in a stable grasp. They use the calculus of variations (with high computational cost) to obtain optimal trajectories and actuator forces and torques for object manipulation in the presence of obstacles. Their planning scheme only plans for the pushing action, assuming that robots have already grasped the object and do not need to plan for the grasping position.

A few model-based hybrid manipulation controllers have been introduced (\cite{woodruff2017planning,hogan2017reactive,hogan2016feedback}). The control strategies presented in the aforementioned papers are applied to systems with a priori knowledge of the contact mode sequencing or offline determination of optimal mode sequences. In \cite{hogan2017reactive}, MPC is used to find an optimal sequence of robot motions to achieve the desired object motion. \cite{pajarinen2017hybrid} solves the problem of finding an optimal sequence of hybrid controls under uncertainty using differential dynamic programming and incorporating discrete actions inside DDP.

We introduce an online MPC-based manipulation planning framework in which we plan for both the object and the robot together to reach a desired configuration for the object. We leverage the learned dynamics model without any prior knowledge of contact mode sequences. We use replanning to compensate for our modeling errors.

\section*{Problem Formulation and Framework}
The problem of mobile manipulation of legged objects can be broken into four main sub-problems:

\begin{myenumerate}
\item Moving the robot from the initial position to the object grasping position.

\item Grasping the object's leg.

\item Manipulation of the object to reach the desired configuration.

\item Re-positioning to avoid infeasible configurations.

\end{myenumerate}

The first sub-problem is essentially a collision-free mobile robot motion planning problem, conditioned on a grasping goal. We define the planning problem as an optimization to obtain an optimal path from the initial configuration to the desired configuration. For the grasping sub-problem, to avoid complications due to the robot's limitations, we use a predefined sequence of actions. The motion planning and manipulation planning sub-problems share the same core optimization structure, with some additional constraints for the manipulation planning that incorporate the dynamics of the manipulated object.

The re-positioning sub-problem covers the cases where we need to prevent the robot from going into infeasible configurations. For example, in our experiments, the robot can become stuck between two legs of the object, which is an undesirable situation. In that configuration, the object's motion causes disturbing forces on the robot that affect the robot's motion. Although the re-positioning sub-problem is specific to the design, other robots would also have similar issues with moving into configurations that require re-positioning to avoid undesirable configurations or the performance limits. We discuss the details of each sub-problem in the ``Mobile  Manipulation Framework'' section below.

The focus of this paper is on the manipulation planning algorithm which is defined as controlling a dynamical system 
\begin{equation}
\dot{x} = f(x,u)
\end{equation}
such that it takes the system to a desired state.
For solving this problem, we consider the discrete form of the dynamics which is an approximation of real dynamics of the system:
\begin{equation}
x_{t+1} = \hat{f}(x_t,u_t)
\label{dynamic_approximated}
\end{equation}
With $x_t$ and $u_t$ representing system state and control input, respectively, at time $t$.

To control the above system, we need three main components: (1) A robot model, which defines the kinematics of the robot based on a given control input ($u$). (2) An object dynamics model ($\hat{f}$) to estimate the motion of objects given an applied force. (3) A manipulation planner to plan a feasible sequence of control inputs based on the robot kinematics model and object dynamics model to reach the desired configuration.

In this paper, we use an MPC approach to find the desired control input. At each time step, we find the optimal control sequence $\mathbf{u}=\{u_{t+1},...,u_{t+H}\}$ for a limited horizon $H$ following the approximated dynamics model $\hat{f}$ resulting in a sequence of system states $\mathbf{x}=\{x_{t,1}, x_{t,2},...,x_{t,H}\}$. We apply the first control input in the sequence $u_{t+1}$, and replan for the next step, until the system reaches a state in the goal set. Through this replanning framework, we desire to minimize an optimal control objective, which in a mobile manipulation problem consists of two main components of the system, the robot and the object. In other words, we need to find the state control sequence which solves
\begin{equation}
\min_{\mathbf{x},\mathbf{u}}\; c(\mathbf{x},\mathbf{u})
\end{equation}
with respect to the constraints of the problem which includes obstacle avoidance, as well as kinematics and dynamics constraints. Once this optimization is solved, we extract the first control input and apply it to the real system $f$. Since, at each time step, we replan based on the current state $x_t$, the entire controller behaves as a state-feedback controller and is able to compensate for local model and perception errors. The objective function is defined as the performance of the robot and the object:
\begin{equation}
c(\mathbf{x},\mathbf{u}) = c_R(\mathbf{x},\mathbf{u}) + c_O(\mathbf{x},\mathbf{u})
\end{equation}
$c_R$ represents all the costs associated with robot's motion and $c_O$ is defined as the overall cost for the object's motion. Each cost function consists of a cost over the entire horizon $c_P$, and a terminal cost considering the last state in the horizon $c_T$. 
\begin{equation}
\begin{split}
c_i(\mathbf{x},\mathbf{u}) = \Sigma_{h=0}^{H-1} c_{i,P}(x_{t,h},u_{t,h}) + c_{i,T}(x_{t,H},u_{t,H}),\\
i\in\{R,O\}
\end{split}
\end{equation}
We will define these specific costs for the application of legged objects manipulation.
In the following sections, we discuss the details of how we find an approximation for the system dynamics and the manipulation planning framework using MPC.

\section*{Robot Model}
For the experimental studies, we use the mobile manipulator introduced in (\mbox{\cite{novin2018dynamic}}), which is a 3-finger gripper mounted on an iRobot Create2. The gripper has a single actuator to move all three fingers and a low-level current (torque) controller to control the torque in different gripper modes (Open, Close, Stay-Put).

The iRobot Create2 is a nonholonomic mobile robot with differentially-driven wheels controlled with a low-level velocity controller. The inputs to the velocity controller are the linear velocity and the radius of motion curvature, and the output is the linear velocities of the wheels.

\begin{figure}[t!]
\center
\includegraphics[width=7.5cm]{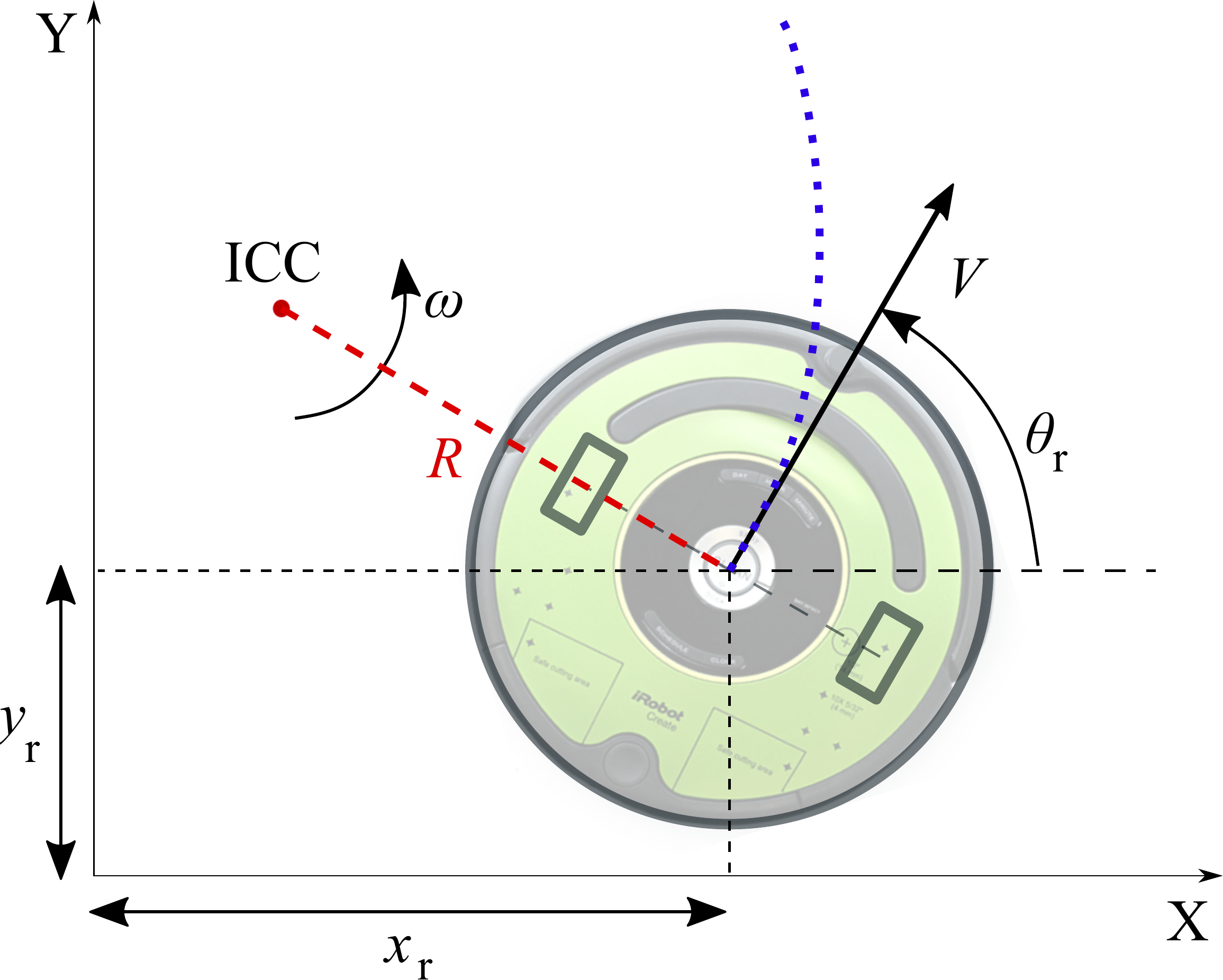}
\caption{Kinematics model of the mobile robot.}
\label{robot_model}
\end{figure}

Denoting the robot's pose by \mbox{$\boldsymbol{\xi}_r = [x_r, y_r, \theta_r]^T$}, linear and angular velocities are defined as \mbox{$v_r = \sqrt{\dot{x}_r^2 +\dot{y}_r^2}$} and \mbox{$\omega_r = \dot{\theta}_r$}, respectively (\mbox{Fig.~\ref{robot_model}}). When the robot is moving on a curve, there is a center of motion curve at that moment, known as the Instantaneous Center of Curvature (or ICC). The distance of the robot center to ICC which is called the radius of curvature is obtained as:

\begin{equation}
   R = \frac{v}{w}
\end{equation}

The maximum speed of the robot is $0.5$ m/s in both forward and reverse directions, and rotations can be performed up to $2$m curvature radius (\mbox{\cite{iRobot2018}}). This model defines the primary constraints we consider in the manipulation optimization below to ensure physically realizable solutions.

\section{Dynamics Models of Objects}

\begin{figure}[t!]
\center
\includegraphics[width=6.5cm]{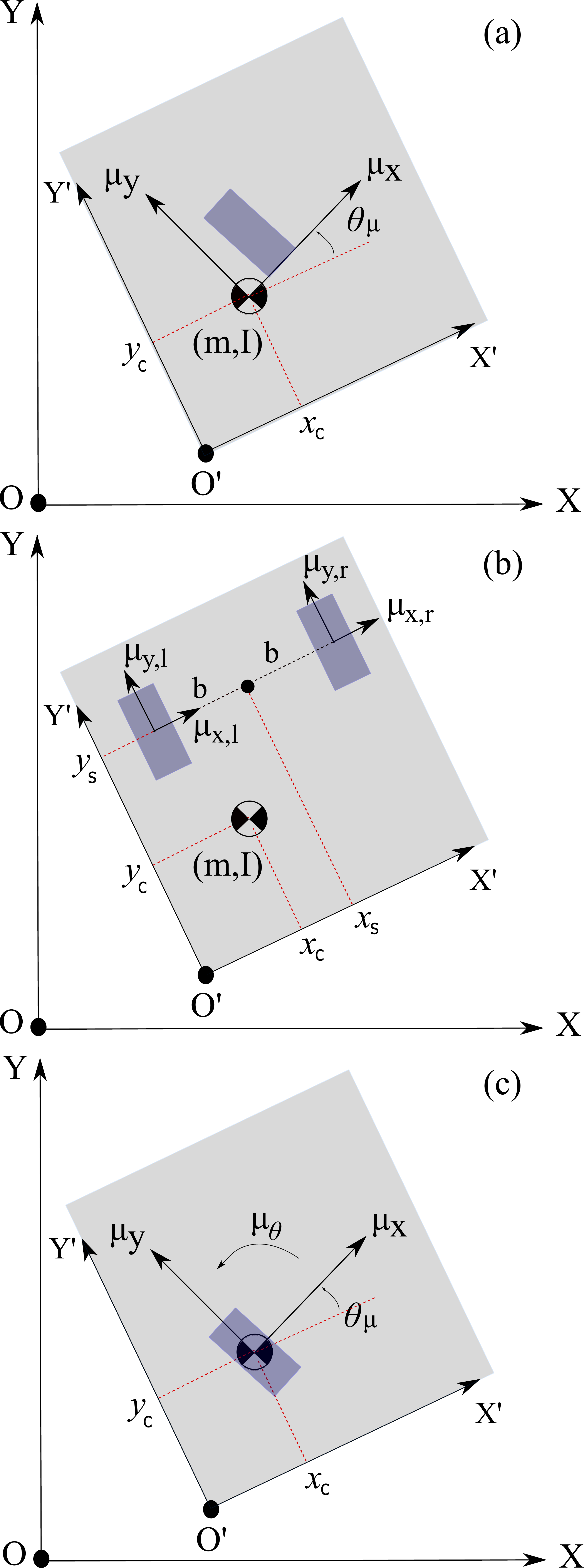}
\caption{Three different dynamics models with inertia and friction parameters used in Bayesian regression. (a)~A point mass on a wheel model. (b)~2-wheel model. (c)~Friction-only model.}
\label{model}
\end{figure}

For object motion estimation, we prefer learning the dynamics model since our task (i.e., which object to move) is not defined a priori. We use Bayesian regression to predict dynamic parameters from observed motion and force data. For this purpose, we consider three different models: (1) a simple model of a point mass on a wheel, (2) a 2-wheel model, (3) a friction-only model. These models are shown in Fig.~\ref{model}. The method we use here is adopted from \cite{scholz2015learning}. The obtained model will provide us with a probabilistic estimate of the object's dynamic parameters for a given model. 

Input data includes exerted force to the leg and the resulting object motion. We only apply force and assume the applied torque is zero. The reason to avoid collecting torque data is to develop a simple and general model with a small dataset. However, the algorithm could be implemented with a torque sensor as well. 

We only consider planar parameters since the objects of interest will only slide or roll on the floor. Dynamic parameters for a planar model include inertia and friction. For the first two models, inertia requires four parameters for planar manipulation: one for mass, one for inertia in the $\mathrm{XY}$ plane, and two for the center of mass position. The difference between these two models comes from friction. For the point mass on a wheel, we define two friction coefficients $\mu_x$ and $\mu_y$ in the $\mathrm{X}$ and $\mathrm{Y}$ directions, respectively, to define the anisotropic friction and $\theta_{\mu}$ for the wheel orientation. In this case, the model parameter vector is:
\begin{equation}
\mathbf{\Phi}_1 := <m, I, x_c, y_c, \mu_x, \mu_y, \theta_{\mu}>
\end{equation}

However, for the second model, we have two wheels, resulting in four friction coefficients, $(\mu_{x,r}, \mu_{y,r})$ for the right wheel and $(\mu_{x,l}, \mu_{y,l})$ for the left one. We assume the orientation of the wheels are known. Another set of parameters define the position of wheels which includes a center of wheel shaft position ($x_s$, $y_s$) and the distance between two wheels ($2b$). The parameter vector for this model is:
\begin{equation}
\mathbf{\Phi}_2 := <m, I, x_c, y_c, \mu_{x,r}, \mu_{y,r}, \mu_{x,l}, \mu_{y,l},x_s, y_s, b>
\end{equation}

In the third model, we investigate the effect of inertia by only considering the friction parameters for the 1-wheel model. For this, we introduce a new friction term $\mu_{\theta}$ to represent resistance to rotation, resulting in four friction coefficients in total. The parameter vector for this model is:
\begin{equation}
\mathbf{\Phi}_3 := <x_c, y_c, \mu_x, \mu_y, \mu_{\theta}, \theta_{\mu}>
\end{equation}

We use Bayesian regression to find these parameters, which are presented as random variables from a prior probability distribution in the model. Then, we find the conditional probability of possible values of these random variables based on the given observation. Since the posterior distribution cannot be reasonably obtained by direct computation, we use a Markov Chain Monte Carlo (MCMC) method to sample from the distribution~(\cite{bernardo2001bayesian}). We define physics-based prior distributions for dynamic parameters and present them as a truncated normal distribution with mean value $\mathbf{\mu}$ and standard deviation $\mathbf{\sigma}$ since all these parameters have lower and upper bounds $(\mathbf{l_l},\mathbf{l_u})$:
\begin{equation*}
\mathbf{\Phi}\sim N_t(\mathbf{\mu}, \mathbf{\sigma}, \mathbf{l}_l, \mathbf{l}_u)
\end{equation*}

Next, we derive the dynamics equations. There are various friction models including viscous and coulomb frictions (\mbox{\cite{olsson1998friction}}). However, due to the low velocities of manipulation in our experiments, we only considered the viscous friction. Following (\mbox{\cite{scholz2015learning}}), we used anisotropic friction constraint to simplify the friction model to two orthogonal friction coefficients.

To obtain the friction force for each wheel in the models, we compute the velocity of the wheel in the wheel frame, find the friction force components, and finally convert it back to the world frame. Using the second model, the wheel frame and object frame are the same as the orientation of the wheels are fixed. The velocity of the wheel in the wheel frame is computed based on the center of mass velocity as:
\begin{equation}
\mathbf{v}_{w}=\mathbf{R}'^{-1}\mathbf{R}^{-1}\bigg(\begin{bmatrix}\dot{x}\\\dot{y}\end{bmatrix} + \begin{bmatrix}0 & -\dot{\theta}\\\dot{\theta} & 0\end{bmatrix}\mathbf{R}\begin{bmatrix}x_w\\y_w\end{bmatrix}    \bigg)
\end{equation}

which results in the wheel friction force ($\mathbf{F}_{w}$):
\begin{equation}
\mathbf{F}_{w}= \mathbf{R}'\mathbf{R}\begin{bmatrix}\mu_{x} & 0\\0 & \mu_{y}\end{bmatrix}\mathbf{v}_{w}
\end{equation}
In the above, $\mathbf{R}$ is a rotation matrix from the world frame to the object frame and $\mathbf{R}'$ is a rotation matrix from the object frame to the wheel frame. $[\dot{x},\dot{y},\dot{\theta}]$ represent the object's planar velocity and $[x_w,y_w]$ are position of the wheel in the object frame. In the 1-wheel model, this is the same as the center of mass position. However, in the second model, the position of the wheels are defined using the center of wheel shaft $(x_s,y_s)$ and the shaft length $(b)$:
\begin{equation}
x_w = x_s \pm b, \quad y_w = y_s
\end{equation}
The total force/torque $\mathbf{F}_T$ is the sum of all friction forces/torque $(\mathbf{F}_{w,i},\boldsymbol{\tau}_{w,i})$ and the input force/torque from the robot $(\mathbf{F}_r, \mathbf{\tau}_r)$ which results in the following dynamics:
\begin{equation}
\mathbf{F}_T = \begin{bmatrix}\mathbf{F}\\\boldsymbol{\tau}\end{bmatrix}=\begin{bmatrix}\mathbf{F}_r+\Sigma \mathbf{F}_{w,i}\\\mathbf{R}_r\times \mathbf{F}_r+\Sigma (\mathbf{R}_{w,i}\times \mathbf{F}_{w,i})\end{bmatrix}
\end{equation}
\begin{equation}
\ddot{\mathbf{x}}=\mathbf{I}^{-1}\mathbf{F}_T
\label{dynamic}
\end{equation}
where $I$ denotes the object's inertia matrix and $\ddot{\mathbf{x}}$ is the object's acceleration. Taking the input as the applied force by the robot ($\mathbf{u}=F_r$) and reforming Eq.~\ref{dynamic_approximated} as a discrete time integration over time steps $\delta t_i$ and additive Gaussian noise $\epsilon_i$ with zero mean and variance $\sigma^2$ results in:
\begin{equation}
x_{i+1}=\hat{f}(x_i,u_i,\delta t_i,\mathbf{\Phi})+\epsilon_i
\end{equation}
which defines our Bayesian regression model. Both input variables and output noise include uncertainty. We find the probability of dynamic parameters and output noise using the input dataset $\mathcal{D}=\{\mathbf{x},\mathbf{u}\}$ and Bayes theorem:
\begin{equation}
P(\mathbf{\Phi},\sigma|\mathcal{D})\propto P(\mathcal{D}|\mathbf{\Phi},\sigma)P(\mathbf{\Phi})P(\sigma)
\end{equation}

\section{Mobile Manipulation Framework}

\begin{figure}[t!]
\center
\includegraphics[width=8.5cm]{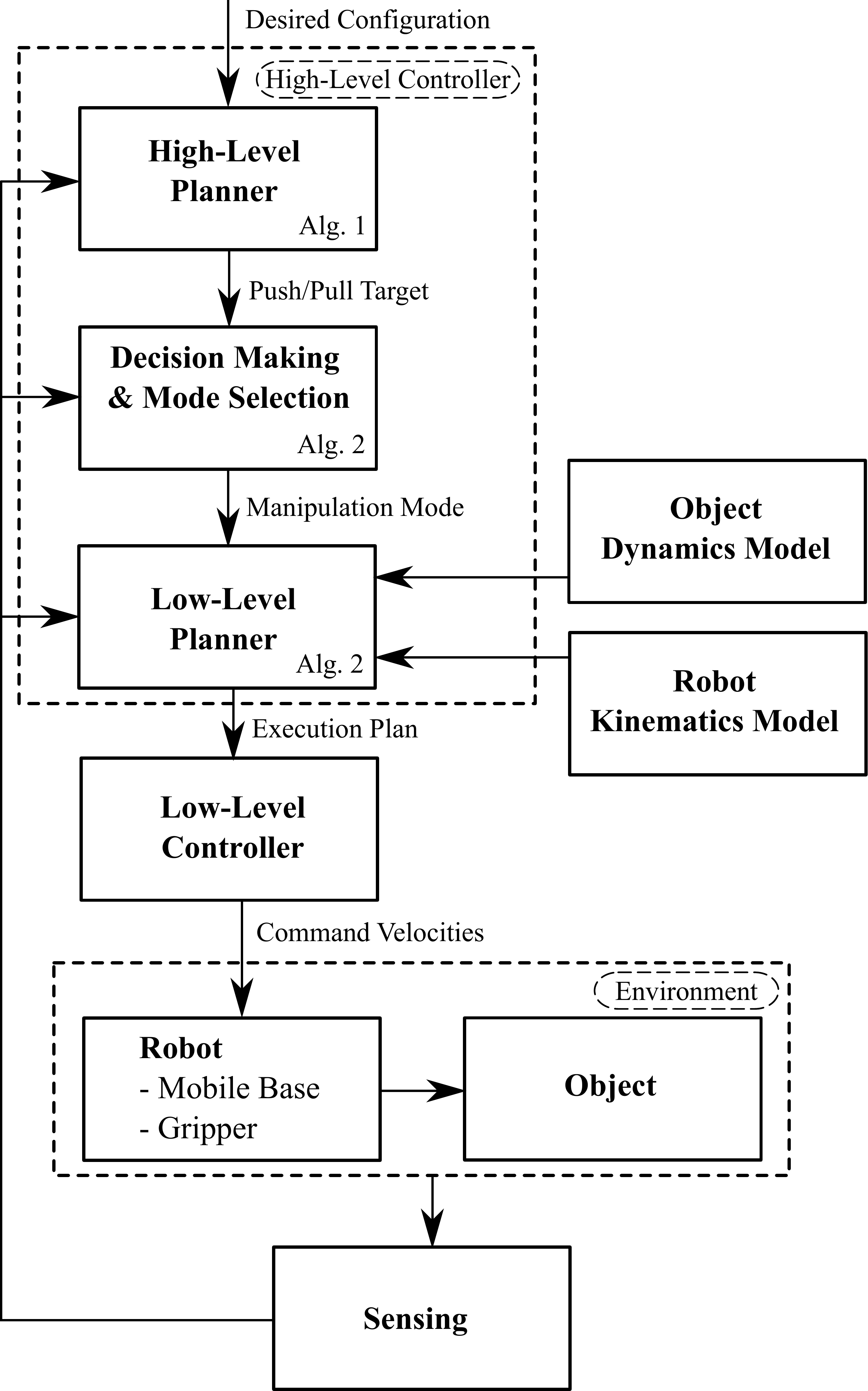}
\caption{The proposed structure of the mobile manipulation framework and connection of various components.}
\label{framework}
\end{figure}

In this section, we discuss various parts of our manipulation planning framework. Figure \mbox{\ref{framework}} presents the overall structure of our proposed framework. As discussed earlier, the manipulation planning algorithm is divided into four major modes: (1) motion planning from the robot's initial position to the object grasping position, (2) grasping mode to grasp the object's leg, (3) manipulation planning to move the object to the desired configuration, and (4) re-positioning mode to release and re-grasp from a better direction. To define the MPC objective function, we have to consider all four modes. 

In the first two modes, the object is stationary, so we only define the cost for the robot motion. In the third mode, both the robot and the object are involved; however, when the object is grasped, it moves with the robot. Thus, minimizing the cost for the robot or the object also minimizes the entire system's cost. 

For this mode, since the smoothness of the object path is more important for us, we define the cost as the path length of the object. At each time step, we assume that the robot is not changing legs during the current horizon. However, we know that this may not necessarily be true because the robot can get stuck between two legs of the object. In this case, replanning and repositioning are necessary. 

 To find the cost of repositioning, after planning optimization, we simulate the planned trajectory from optimization with the approximate dynamics. Then, we find the cost by counting the number of repositioning actions needed for that plan and add it to the total cost:
\begin{equation}
c_T = c_{motion} + c_{grasp} + c_{manipulation} + c_{reposition}
\label{total_cost}
\end{equation}

In the following, we discuss how we calculate each of these costs and choose the optimal leg for manipulation. At each time step, we calculate the total cost for each leg of the object and choose the one with the minimum total cost. Algorithm~\ref{manipulation} provides the pseudocode of the high-level planner which is also referred to as function (``opt") in Algorithm~\mbox{\ref{high_level}}. The inputs to this function are the target object's current and goal states ($\boldsymbol{\xi}_O,\boldsymbol{\xi}_O^g$), the robot's current state ($\boldsymbol{\xi}_r$), and all other objects which are considered as obstacles ($\mathcal{O}$). 

First, we find a manipulation plan for each leg of the object, assuming that the robot has grasped that leg~($c_{manipulation}$). However, since the dynamic constraints are not linear and cannot be used in the convex optimization directly, we find a nominal trajectory based on the object's current and desired state (Line 4) and linearize the dynamic constraints over the nominal trajectory. To find the nominal trajectory, we use a simple version of a path planning problem, with obstacle avoidance but without dynamics constraints, for a horizon length that is the same as the main problem.

\setlength{\textfloatsep}{2pt}
\begin{algorithm}[!t]
\DontPrintSemicolon
\SetAlgoLined
\SetNlSty{texttttt}{(}{)}
\SetKwInOut{Input}{input}
\SetKwInOut{Output}{output}
\KwResult{$\pi^*$} 
\Input{$\boldsymbol{\xi}_o, \boldsymbol{\xi}_o^g, \boldsymbol{\xi}_r, \mathcal{O}$}
$c_{\min}$ $\leftarrow$ $\infty$\;
$\pi^*$ $\leftarrow$ $\emptyset$\;
\For { $l$ $\mathrm{in}$ $\mathbf{legs}$}{
$\mathcal{T}_n$ $\leftarrow$ $\mathrm{nominal\_traj}$($\boldsymbol{\xi}_o, \boldsymbol{\xi}_o^g, \mathcal{O}$) \;
$c_{manipulation}$, $\pi$ $\leftarrow$ $\mathrm{opt\_manipulate}$($\boldsymbol{\xi}_o, \boldsymbol{\xi}_o^g, \mathcal{O}, \mathcal{T}_n$)\;
$c_{motion}$ $\leftarrow$ $\mathrm{opt\_motion}$($\boldsymbol{\xi}_r, \pi,\mathcal{O}$)\;
$c_{reposition}$ $\leftarrow$ $\mathrm{simulate}$($\pi$)\;
$c_{grasp}$ $\leftarrow$ $\mathrm{change\_leg}$($l_c, l$)\;
$c_T$ $\leftarrow$ $c_{manipulation}$ + $c_{motion}$ + $c_{reposition}$ + $c_{grasp}$\;
\If{$c_T < c_{\min}$}{
$c_{\min}$ $\leftarrow$ $c_T$\;
$\pi^*$ $\leftarrow$ $\pi$\;
    }
    }
return $\pi^*$\;
\caption{High-level planner used in ``opt"}
\label{manipulation}
\end{algorithm}

Based on the obtained manipulation plan, we find a grasping goal for the robot and find the cost of the robot motion to get to the grasping goal ($c_{motion}$), which is calculated in the "opt\_motion" function. If there is a need for a new grasp, we add grasping cost ($c_{grasp}$), which is the cost of releasing the current leg and grasping another one. Otherwise, if the plan is for the current grasp, we assign zero grasping cost. To find the cost for repositioning, since this is not directly included in the optimization, we simulate the entire manipulation plan and count the number of repositioning actions required for that plan and use that to calculate the cost of repositioning ($c_{reposition}$). Combining all four costs gives us the cost for one leg. We repeat this procedure for all legs and find the minimum cost leg. 

If the minimum cost is infinity, it means a feasible trajectory was not found, so the robot stays put. Otherwise, the ``opt" function in Algorithm~\mbox{\ref{high_level}} returns the optimal plan $\pi^*$. It should be mentioned that we scale all the weights in such a way that the robot avoids changing legs when the object is far from the desired configuration and gradually decrease the weight for grasping to allow the final adjustments of the object's configuration. 

Once the minimum cost plan is chosen, the current mode of the robot is found based on the distance of the robot to the desired leg and also the distance of the object to the desired configuration. Algorithm~\ref{high_level} presents the high-level framework which controls the modes and sends commands to the low-level controller of the robot according to the current planning mode. The inputs to the algorithm are: object current state $\boldsymbol{\xi}_o$, object goal state $\boldsymbol{\xi}_o^g$, robot current state $\boldsymbol{\xi}_r$, set of obstacles $\mathcal{O}$ and dynamics parameters $\boldsymbol{\Phi}$. The goal zone for the object and the robot are defined by $\epsilon_o$ and $\epsilon_r$.

At each time step, while the object is not in the goal region ($||\boldsymbol{\xi}_o-\boldsymbol{\xi}_o^g||>\epsilon_o$), if there is no valid plan or there is any change in the environment, including the object's state, we run the complete planning optimization using the ``opt" function (line 6,7). If it finds a feasible plan, it goes to one of the four modes mentioned above:

\begin{myenumerate}

\item If the robot is far from the object, it is in motion planning mode (lines 22-23), and we use motion planning optimization function ``opt\_motion" to find the optimal plan. 

\item After it reaches the grasping zone ($||\boldsymbol{\xi}_r-\boldsymbol{\xi}_r^g||<\epsilon_r$), it goes into the grasping mode until it successfully grasps the object's leg using a pre-defined grasp plan (lines 19-20). 

\item Then, the manipulation mode starts (line 16-17) and replans the manipulation at each step using the ``opt\_manipulation" function. It will repeat this until the object is in the goal zone. 

\item Whenever the robot is close to getting stuck between two legs of the object, it will go to repositioning mode (lines 13-14) and continue from a new direction (line 2-3). This is only for the situation in which the current leg is still the best leg for manipulation, and only the position of the robot is not favorable.
\end{myenumerate}

\setlength{\textfloatsep}{2pt}
\begin{algorithm}[!t]
\DontPrintSemicolon
\SetAlgoLined
\SetNlSty{texttttt}{(}{)}
\SetKwInOut{Input}{input}
\SetKwInOut{Output}{output}
\KwResult{Next action} 
\Input{$\boldsymbol{\xi}_o, \boldsymbol{\xi}_o^g, \boldsymbol{\xi}_r,\mathcal{O},\mathbf{\Phi}$}
\While{$||\boldsymbol{\xi}_o-\boldsymbol{\xi}_o^g||>\epsilon_o$}{
\uIf{$\mathrm{mode}$ = ``$\mathrm{re\mbox{-} positioning}$"}{
$\pi$ $\leftarrow$ move\_to\_robot\_goal$(\boldsymbol{\xi}_r,\boldsymbol{\xi}_r^g,\mathcal{O})$\;
 }
\uElse{
\uIf{$\mathrm{change}$ $\mathrm{\mathbf{or}}$ $\pi = \emptyset$}{
$\pi,l^*,\boldsymbol{\xi}_r^{g}$ $\leftarrow$ opt$(\boldsymbol{\xi}_o, \boldsymbol{\xi}_o^g, \boldsymbol{\xi}_r,\mathcal{O},\mathbf{\Phi})$\;
$l_c$ $\leftarrow$ find\_current\_leg$(\boldsymbol{\xi}_o, \boldsymbol{\xi}_r)$\;
  }
\uIf{$\pi \neq \emptyset$}{
\uIf{$l_c = l^*$}{
$\mathrm{out\_of\_limits}$ $\leftarrow$ check\_limits$(\boldsymbol{\xi}_o, \boldsymbol{\xi}_r)$\;
\uIf{out\_of\_limits}{
mode $\leftarrow$ ``re-positioning"\;
$\boldsymbol{\xi}_r^g$ $\leftarrow$ find\_new\_goal$(\boldsymbol{\xi}_r, \pi)$\;
  }
\uElse{
mode $\leftarrow$ ``manipulation"\;
$\pi$ $\leftarrow$ opt\_manipulate$(\boldsymbol{\xi}_o, \boldsymbol{\xi}_o^g,\mathcal{O},\mathbf{\Phi})$\;
  }
  }
\uElseIf{$||\boldsymbol{\xi}_r-\boldsymbol{\xi}_r^g||<\epsilon_r$}{
mode $\leftarrow$ ``grasping"\;
$\pi$ $\leftarrow$ grasp\_plan$(\boldsymbol{\xi}_r,\boldsymbol{\xi}_o,\mathcal{O})$\;
  }
\uElse{
mode $\leftarrow$ ``robot motion"\;
$\pi$ $\leftarrow$ opt\_motion$(\boldsymbol{\xi}_r,\boldsymbol{\xi}_r^g,\mathcal{O})$\;
  }
 }
 }
\uIf{$\pi = \emptyset$}{
return StayPut\;
}
\uElse{
return $\pi[0]$\;
}
}
\caption{High Level Controller}
\label{high_level}
\end{algorithm}

At each step, based on the feedback, the robot can decide that continuing with another leg has a less estimated total cost. In that case, it will release the grasped leg and do the motion planning and the grasping parts all over again for the new leg. The cost of this procedure is considered in the total cost; therefore, this will not happen unless the manipulation cost is significantly improved by changing legs, or finishing the task with only one leg is not feasible at all. 

In the next sections, we explain the structures of different modes in the algorithm.

\subsection{Motion and manipulation modes}

The motion planning and manipulation planning modes share the same core optimization structure, with some additional constraints for the manipulation mode to incorporate the dynamics of object manipulation. We define the planning problem as an MPC-based optimization to obtain an optimal path from the initial configuration to the desired configuration. 

We formulate our planner as a mixed-integer convex optimization problem~(\cite{boyd2004convex}), which is defined as a general optimization problem with convex objective function $J(\mathbf{x},\mathbf{u})$ and convex inequality functions $g_i(\mathbf{x},\mathbf{u})$ or piecewise affine equality functions $f_i(\mathbf{x},\mathbf{u})$ as constraints:
\begin{equation*}
\begin{aligned}
& \underset{\mathbf{x}}{\text{min}}
& & J(\mathbf{x},\mathbf{u}) \\
& \text{s.t.}
& & f_i(\mathbf{x},\mathbf{u}) = 0, \; i = 1, \ldots, m\\
& & & g_j(\mathbf{x},\mathbf{u}) < 0, \; j = 1, \ldots, n.
\end{aligned}
\end{equation*}

In the following, we provide details on all components of the optimization problem used both in motion planning for the robot and manipulation planning of the object.
\\

\text{\textbf{Cost function:}}

Since we are using a convex optimization framework, the cost function must have a convex form. Here, we define it as the shortest path cost function with the purpose of finding a smooth trajectory around obstacles and furniture, and avoiding unnecessary motion. At this point, we are not concerned about time because the velocity limits of the robot will not allow for fast motion. However, with a more powerful robot, having a combination of minimum time and the shortest path would be a better option. By considering a control horizon with length of $H$, we can write the cost function for the shortest path:
\begin{gather}
c = \omega_1\sum_{h=0}^{H-1}\boldsymbol\delta_P + \omega_2\boldsymbol\delta_T
\end{gather}

where $\boldsymbol{\delta}_P=||\boldsymbol{\xi}_{t,h+1}-\boldsymbol{\xi}_{t,h}||_2^2$ is the change in the object's state between two steps in the horizon and $\boldsymbol{\delta}_T=||\boldsymbol{\xi}_{t,H}-\boldsymbol{\xi}^g||_2^2$ is the terminal cost and shows the difference between the final state in the horizon and the goal state. From now on, for better readability, we drop $t$ in the state notation and use subscription only to show the step in the horizon for a single
receding-horizon iteration. $(\omega_1,\omega_2)$ are weight matrices to adjust based on the importance of each term in the cost function. 

It should be mentioned that in the manipulation problem, we define the optimization based on the object's state. However, for the motion planning mode, which only includes the robot's motion, we use the robot's states to define the cost function. Moreover, in the manipulation planning, we use different values for the weights on position and orientation costs. 

We use lower ratios of orientation cost over position cost when the object is far from the goal state and increase it as the object gets closer to the goal state. This is mainly because, in our application, the object's orientation in the middle of the trajectory is not as crucial as getting the object to the goal position. So, here $(\omega_1,\omega_2)$ change overtime during one task.
\\

\text{\textbf{Obstacle avoidance:}}

The obstacles are written as equivalent surrounding convex forms. Therefore, each obstacle is approximated by a polygonal shape. Although this is slightly more conservative than a point cloud, this assumption does not have a dramatic effect on the optimality of the solution since most of the objects in medical environments are box-shaped like hospital beds or chairs. Polygon shapes are defined as the intersection of a series of half spaces: 
\begin{equation}
\mathcal{O} : \{\boldsymbol{\xi}|\mathbf{A}\boldsymbol{\xi}<\mathbf{b}\}
\end{equation} 
The point $\boldsymbol\xi$ is outside of shape $\boldmath{O}$ with $m$ number of sides if at least one of the $\mathbf{A}\boldsymbol{\xi}<\mathbf{b}$ inequalities is satisfied:
\begin{gather}
\mathbf{A}\boldsymbol{\xi}<\mathbf{b}+(\mathbf{v}-1)M, \quad\quad
\sum_{i=1}^m v_i\geq 1
\label{obs}
\end{gather} 
where $\mathbf{v}$ is a vector of binary variables and $M$ is a large constant value used in the Big-M method~(\cite{richards2005mixed}). Equation~\ref{obs} ensures that at least one element of the vector $\mathbf{v}$ equals to $1$, so point $\boldsymbol\xi$ would be out of the polygonal obstacle.
\\

\text{\textbf{Kinematics constraints:}}

The kinematic constraints include initial state constraint and velocity/acceleration limits of the objects or the robot. Considering equal timesteps $dT$, these constraints can be formulated as below:
\begin{gather}
\boldsymbol{\xi}_0 = \boldsymbol{\xi}_s\\
|\frac{\boldsymbol{\xi}_{h+1}}-\boldsymbol{\xi}_{h}{dT}| \leq \dot{\boldsymbol\xi}_{max} , \; h = 0, \ldots, H
\end{gather}

There is also a limitation on the radius of curvature for the robot, meaning that the angular velocity is either zero, or follows a restricted relationship with the linear velocity:
\begin{gather}
|\frac{\mathbf{v}_r}{\omega_r}| \leq R_{max}\quad \mathrm{or}\quad  \omega_r = 0 
\end{gather}
In the above, $v_r$ and $w_r$ are linear and angular velocities of the robot, and $R_{max}$ denotes the maximum allowed radius of curvature for the robot's motion.
\\

\text{\textbf{Dynamic constraints:}}

Finally, a set of constraints that change the motion planning problem to the manipulation problem is dynamics constraints. Dynamic constraints play the main role in the optimization problem for manipulation and connect the applied force to the object's motion. Dynamic equations, which result from the dynamics learning discussed earlier, define the relationship between the applied force by the robot and the resulting trajectory of the object. 

These constraints are non-linear and should be linearized before we can use them in the convex optimization problem. We do this by finding an approximated nominal trajectory and using it to linearize the dynamics equations. This would add errors to the solution, but implementing it in a receding horizon framework will compensate for it. 
\begin{gather}
\boldsymbol{\xi}_{h+1} = \hat{f}(\boldsymbol{\xi}_h,\mathbf{u}_h) , \quad h = 0, \ldots, H
\end{gather}

Another part of dynamics constraints are the limitations on the magnitude and direction of force.
\begin{gather}
|\mathbf{u}_h| \leq \mathbf{u}_{max} , \quad h = 0, \ldots, H
\end{gather}

To conclude, the whole optimization problem is provided below, used in both the opt-manipulate and opt-motion functions.
\\

\begin{flalign*}
\underset{\mathbf{u}}{\text{min}}
\quad&c = \omega_1\sum_{h=0}^{H-1}||\boldsymbol{\xi}_{h+1}-\mathbf{\xi}_{h}||_2^2 + \omega_2||\boldsymbol{\xi}_H-\boldsymbol{\xi}^g||_2^2\\
 \text{s.t.} \quad&\boldsymbol{\xi}_0 = \boldsymbol{\xi}_s \\
&|\frac{\boldsymbol{\xi}_{h+1}-\boldsymbol{\xi}_{h}}{dT}| \leq \dot{\boldsymbol\xi}_{max}, \quad\qquad\qquad\quad\forall h \in \{0,.., H\}\\
& |\frac{\mathbf{v}_{r,h}}{\omega_{r,h}}| \leq R_{max}\quad \mathrm{or}\quad  \omega_{r,h} = 0
, \quad\quad\forall h \in \{0,.., H\}\\
&  \boldsymbol{\xi}_{h+1} = \hat{f}(\boldsymbol{\xi}_{h},\mathbf{u}_h),  \qquad\qquad\qquad\quad\;\forall h \in \{0,.., H\}\\
& |\mathbf{u}_h| \leq \mathbf{u}_{max}, \qquad\qquad\qquad\quad\quad\quad\;\;\forall h \in \{0,.., H\}\\
& \mathbf{A}_{\mathcal{O}}\boldsymbol{\xi}_{p,h}>\mathbf{b}_{\mathcal{O}}+(\mathbf{v}_h-1)M, \quad\quad\quad\forall h \in \{0,.., H\}\\
& \qquad\qquad\qquad\qquad\qquad\qquad\quad\qquad\;\forall \mathcal{O} \in \mathcal{S}_\mathcal{O}\\
& \sum_{i=1}^m v_{i,h}\geq 1, \qquad\qquad\qquad\qquad\qquad \forall h \in \{0,.., H\}
\end{flalign*}

In the above, $\boldsymbol{\xi}_{p}$ presents only the position part of the object's state used for obstacle avoidance. Weights in the cost function are adjusted and normalized with regards to the problem scale and experiments. The horizon length is set to a value in the range [10, 20]. 

This optimization problem is coded as a mixed-integer quadratic program and is solved efficiently using the Gurobi numerical optimization package (\cite{Gurobi}). The nominal trajectory mentioned in the previous section is obtained by solving a simplified version of this problem, without the nonlinear dynamics constraints.

\subsection{Grasping mode}

For the grasping mode, we use a pre-defined motion primitive to guarantee that the robot can grasp the desired leg of the object. This is mainly because of the robot's limitation in performing fine motion plans. For example, the resolution of rotation is not enough for final adjustments before grasping, or it would take a long time for the robot to perform a lateral transition due to its nonholonomic behavior. 

We change the cost of grasping based on the distance to the goal state. When the robot is far from the goal, we have high re-grasping cost to avoid changing legs as much as possible; but, when the robot gets closer, we want to have the flexibility of using different legs to adjust the object's orientation as desired.

\subsection{Repositioning mode}

In the physical experiments, when the robot is pushing an object, it can sometimes become stuck between two legs of the object because of the object's rotation. This is an undesirable situation because, in that pose, the object will apply some force to the robot, which can affect the robot's motion and prevent it from performing the planned action. 

\begin{figure}[t!]
\center
\includegraphics[width = 8.5cm]{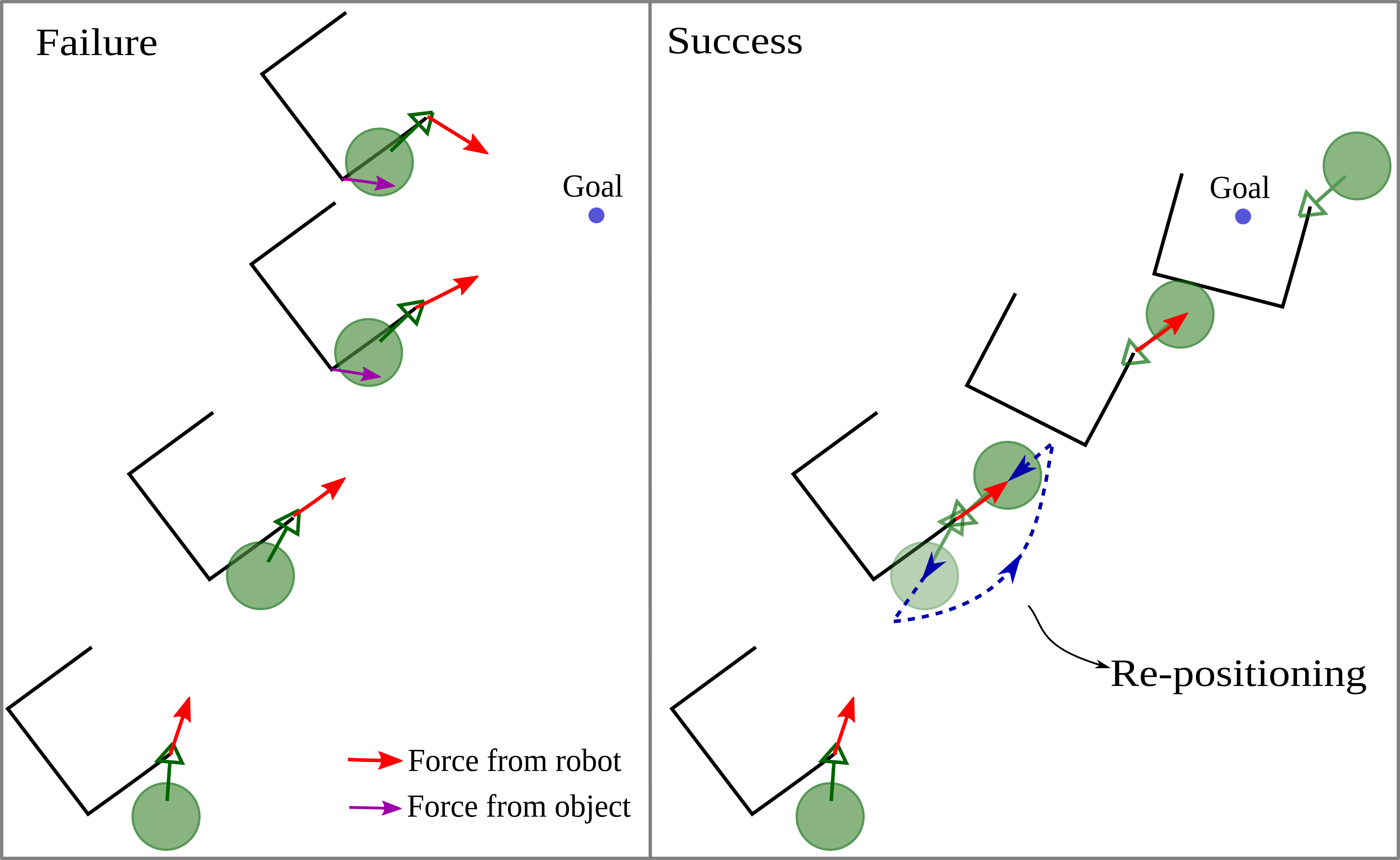}
\caption{Repositioning mode prevents the robot from getting stuck between two legs of the object. On the left figure, when the robot gets stuck and applies force on the object, the other leg also applies some force on the robot. This will change the dynamics of the system and also deviate the robot from it's planned path. On the right figure, by adding the repositioning mode, the robot recovers from this situation and can continue as planned.}
\label{repositioning}
\end{figure}

We avoid this by repositioning whenever the robot gets too close to the object's side, which can lead to getting stuck between legs. After repositioning, the robot will replan and continue manipulation. The repositioning mode includes releasing the leg, moving to the new grasping goal obtained by the desired force direction, and re-grasping. This process and its solution are presented in Fig.~\ref{repositioning}.

The repositioning cost is calculated based on the number of repositioning actions needed to perform the entire plan. We do this as a secondary step by simulating the plan using the dynamics model in a manner the same as the one used in the optimization. Then, we scale it by estimating the average cost of moving from one leg to another.

\begin{figure*}[t!]
\includegraphics[width = 17.5cm]{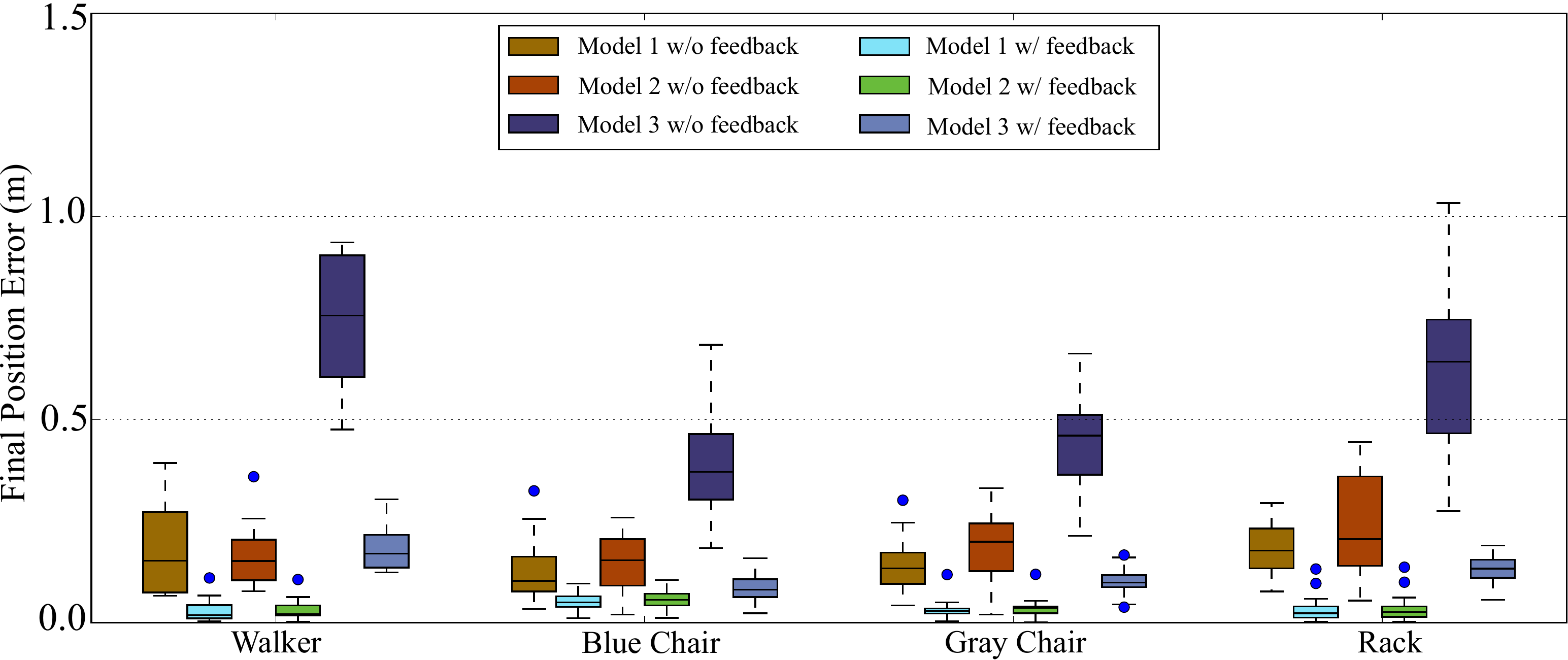}
\caption{Final displacement errors with and without feedback for three different models: (1) Point mass on a wheel model, (2) 2-wheel model, (3) Friction-only model. The displacement error is defined as the difference between predicted and actual displacement at the end of the trajectory.}
\label{rsme}
\end{figure*}

\section{Experiments}

For the experimental studies, we use a low-cost mobile robot based on an iRobot Create2 with a customized 3-finger gripper, introduced in \cite{novin2018dynamic}.

We collect synchronized data for dynamics model learning using a motion capture system with 20 Flex13 cameras (Optitrack, Naturalpoint, Inc.) and a one-directional force sensor mounted on the robot's gripper (Futek Industries). We implement our approach on four different objects; a 2-wheel walker, two 4-wheel chairs, and a 4-wheel rack. The objects are all legged objects, but differ in physical properties, including mass, size, wheel structure, and the leg diameter. The mass of the experimental objects varies from 2 to 5 kilograms. The width and length of bounding boxes of objects are in the range of 0.3m to 0.9m, and the heights of the objects are between 0.8m to 1.2m. Leg diameters vary in the range of 30mm to 50mm.

For each object, we collect data from about 70 short trajectories and divide them to create a 50 element training dataset and a 20 element test dataset. Each trial is about 5-15 seconds and is collected at a 10Hz sampling rate. Force data are filtered by a $6^{th}$ order butterworth filter to smooth the noisy input. In each trial, the robot pushes or pulls one of the object's legs, starting from one of the possible directions. We assume that the robot's gripper acts like a revolute joint and can only apply force, and no torque is applied to the object's leg. It should be mentioned that for some objects, the robot could not grasp all the legs due to the shape of the leg or extra support bars between legs. In those cases, we only collect data from the legs that are graspable by the robot. 

For the Bayesian regression model, we have used the PYMC package~(\cite{patil2010pymc}) in python with 20000 samples running on a Core i7 2.4GHz system. We run model learning for each object using all three models for comparison.

For the simulated manipulation planning experiments, we use the same objects with the learned dynamics models in our simulation setup and run our proposed method to manipulate objects from the initial state to the desired state. We define four different scenarios with various initial states of the object and the robot, the desired state for the object, and room configuration. We run each scenario 50 times for each object and report the success rate, position, orientation error, and average run time. A trial is successful if the robot can take the object to the goal region in less than 3~minutes. Simulations are visualized in Rviz. For evaluation, we compare the performance of our approach with LQR starting with an optimal initial plan obtained from our manipulation optimization.

For the physical experiments, we use a 4-wheel walker, since it is the only object that our robot could apply force in any direction. For the other objects, applying force orthogonal to the gripper's main axis is impossible due to the gripper power limitation. We run our proposed method for two different scenarios, each for five trials, and report the same metrics as simulation experiments. In these experiments, we use the motion capture system as feedback, which adds some delay to the system. 

The collected data for dynamics models learning, source codes of our proposed algorithm, and the optimization parameters can be found at \href{https://sites.google.com/view/mobile-manipulation-planning}{https://sites.google.com/view/mobile-manipulation-planning}.

\section{Results}
In this section, we first discuss final results for the dynamics model learning, showing the errors for all three dynamics models. Then we perform a thorough evaluation of our proposed mobile manipulation planning framework in simulation and physical experiments. 

\begin{figure*}
\center
\includegraphics[width = 16.5cm]{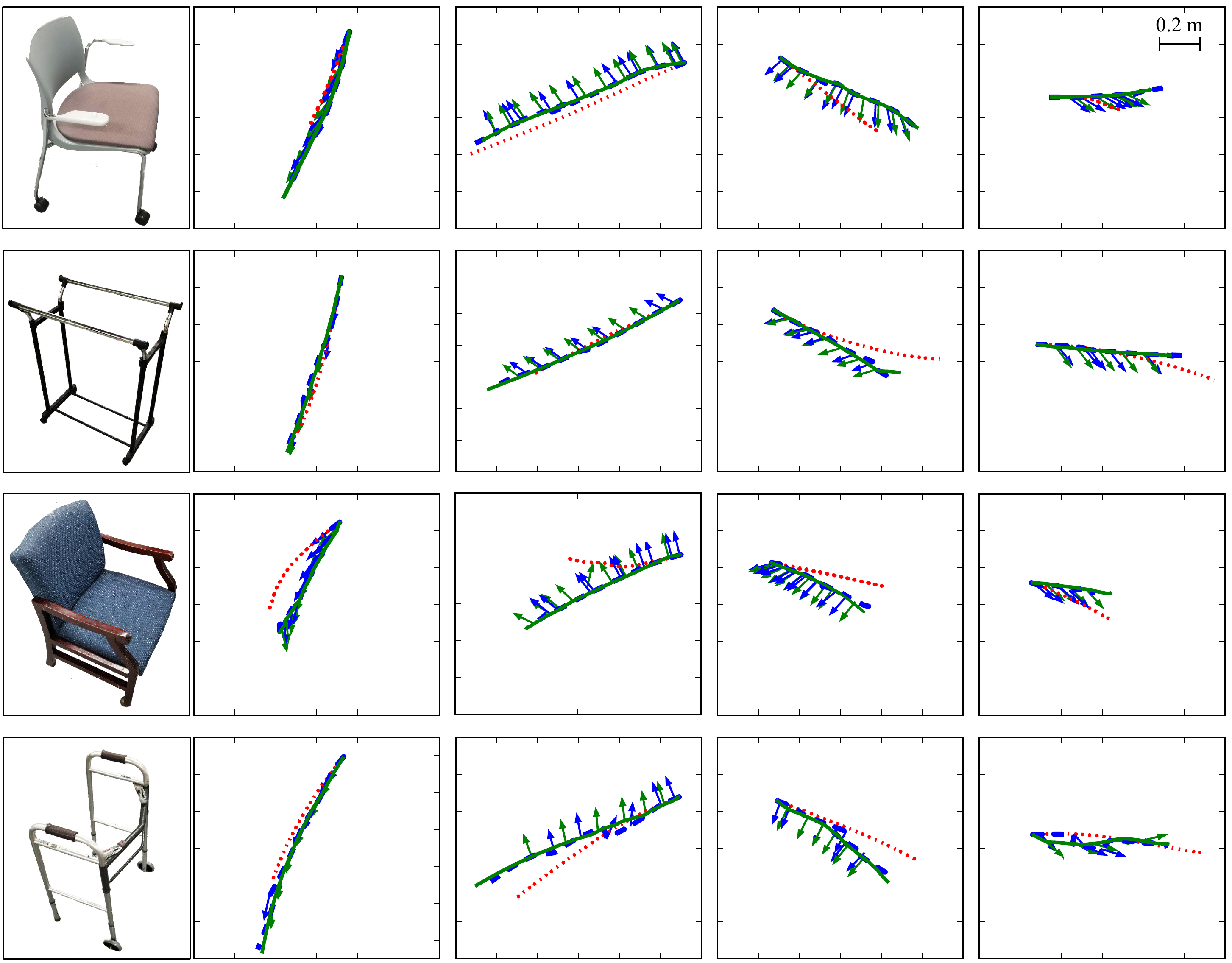}
\caption{Comparison between actual (green solid line) and predicted trajectories of each object, with feedback (blue dashed line) and without feedback (red dotted line) for four different manipulation scenarios. Arrows show the corresponding object orientation. For better visualization the orientation arrow is only shown once for every 10 points in the trajectory. The actual trajectory is from data collected using motion capture. Feedback is every 2 seconds. }
\label{dynamic_results}
\end{figure*}

\begin{figure*}[t!]
\center
\includegraphics[width =17cm]{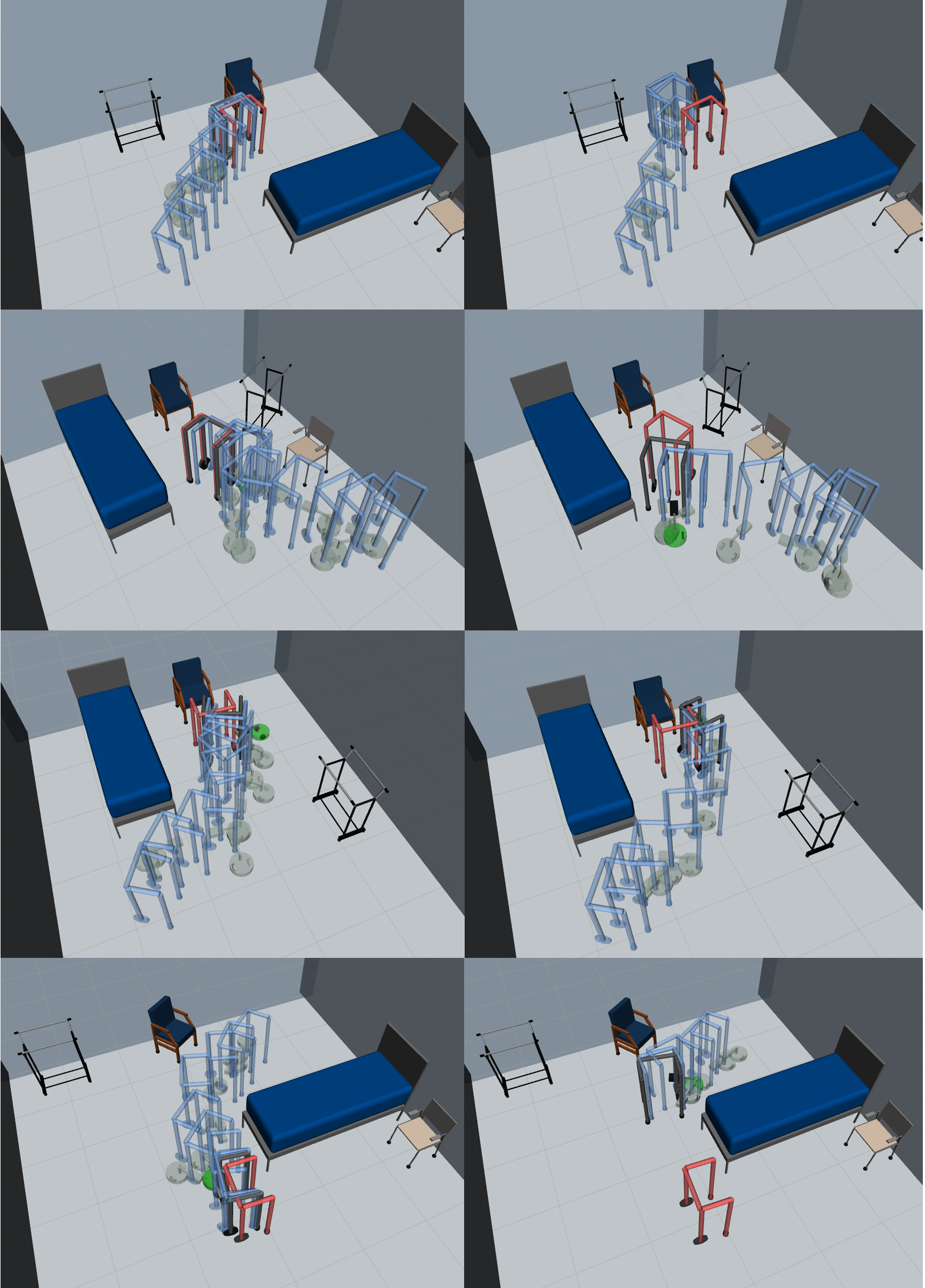}
\caption{Example of simulation results for the walker. The figures on the right show the trajectories using LQR, and on the left are resulting trajectories using the proposed method. The red object shows the goal configuration. In these figures, for better visualization, we only show the manipulation part of planning, excluding motion planning, grasping and repositioning modes.}
\label{sample_results}
\end{figure*}

\begin{figure*}[t!]
\center
\includegraphics[width =17cm]{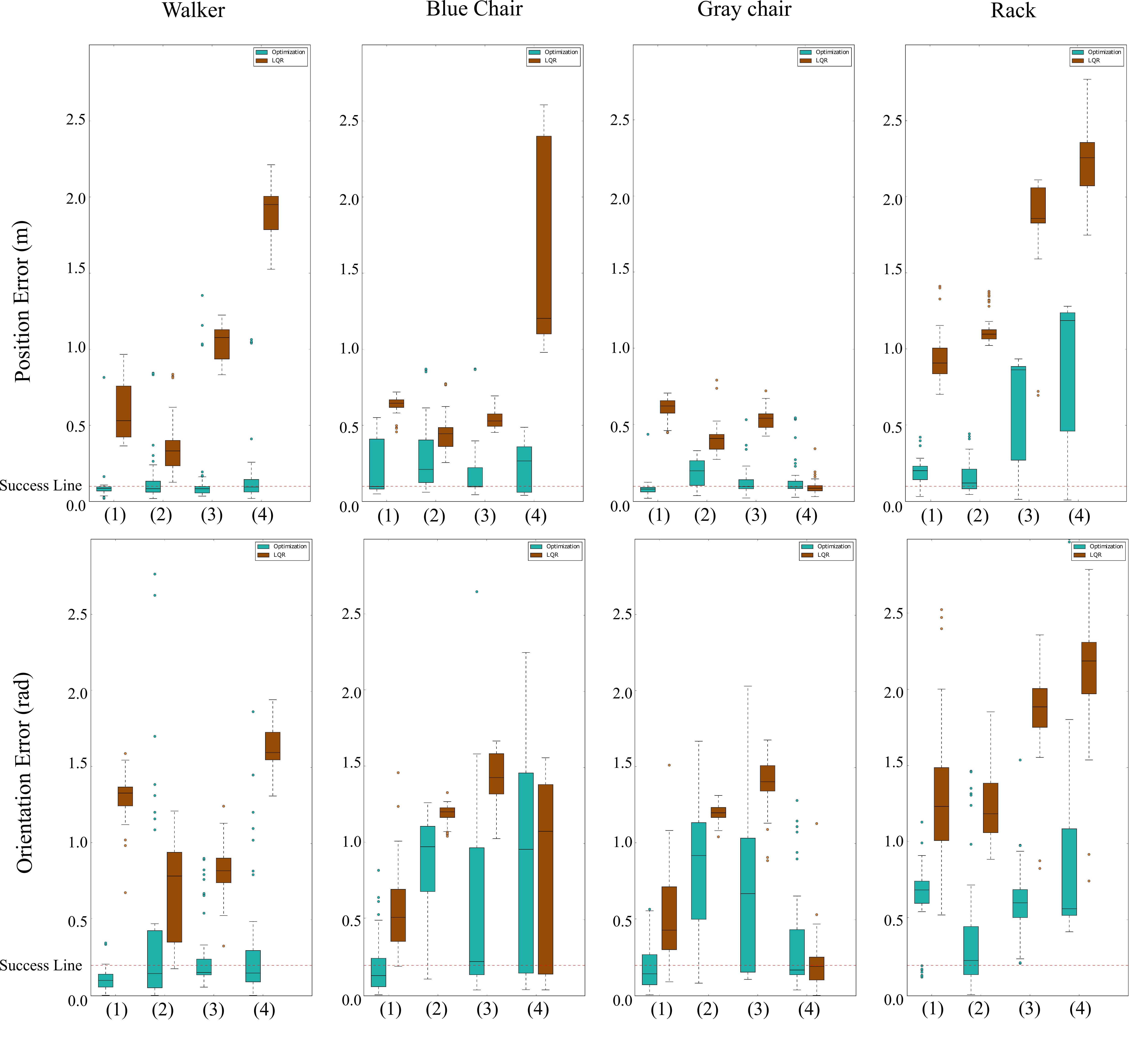}
\vspace{-0.1in}
\caption{Final position and orientation errors for all objects through all tasks. The x-axis shows the task number. We assume that the maximum allowed run time in all trials is 3 minutes, and the system stops after that even if it has not reached the goal region yet. The goal region is defined as a distance less than $10$cm to the goal position with less than $0.2$rad deviation from the goal orientation, which is shown as a red dashed line.}
\label{sim_results}
\end{figure*}

\subsection{Dynamics model learning}

As previously stated, the object's dynamics model is learned using MCMC sampling. On a Core i7 2.4GHz system, it takes about 1 hour to get 20000 samples.

After obtaining an object model based on the training dataset, we tested it on our test dataset containing about 20 trajectory episodes. Each trajectory prediction begins from the actual starting point, and then we only use the real dataset as feedback input every 2 seconds. 

For better evaluation of the model types, for each model, a plot of final displacement errors with and without feedback, which is the difference between predicted and actual displacement at the end of the trajectory, is presented in Fig.~\ref{rsme}. It is shown that including inertia parameters improves prediction significantly. Besides, although the second model works slightly better for the walker, the difference is not significant. As a result, since a more complex model results in higher computational time in optimization, we choose the first model, which is simpler than the second model.

Figure \ref{dynamic_results} provides a comparison between the predicted trajectory without feedback, predicted trajectory with feedback, and the actual trajectory using the first model. As expected, we can see that using feedback helps to stay on the trajectory and eliminate the accumulated error every two seconds; however, it does not change the displacement error much since this is based on the obtained model and noise in the system.

Additionally, we can see that in the far left case, errors get higher as we move towards the end of the trajectory. We believe this is due to forces applied by the gripper fingers, which are not measured in this study. This happens in cases when the force direction is such that gripper fingers apply more force and play a role in influencing dynamics. A better force and torque measurement approach is needed to get more accurate results.

\begin{figure*}[t!]
\center
\includegraphics[width = 17.5cm]{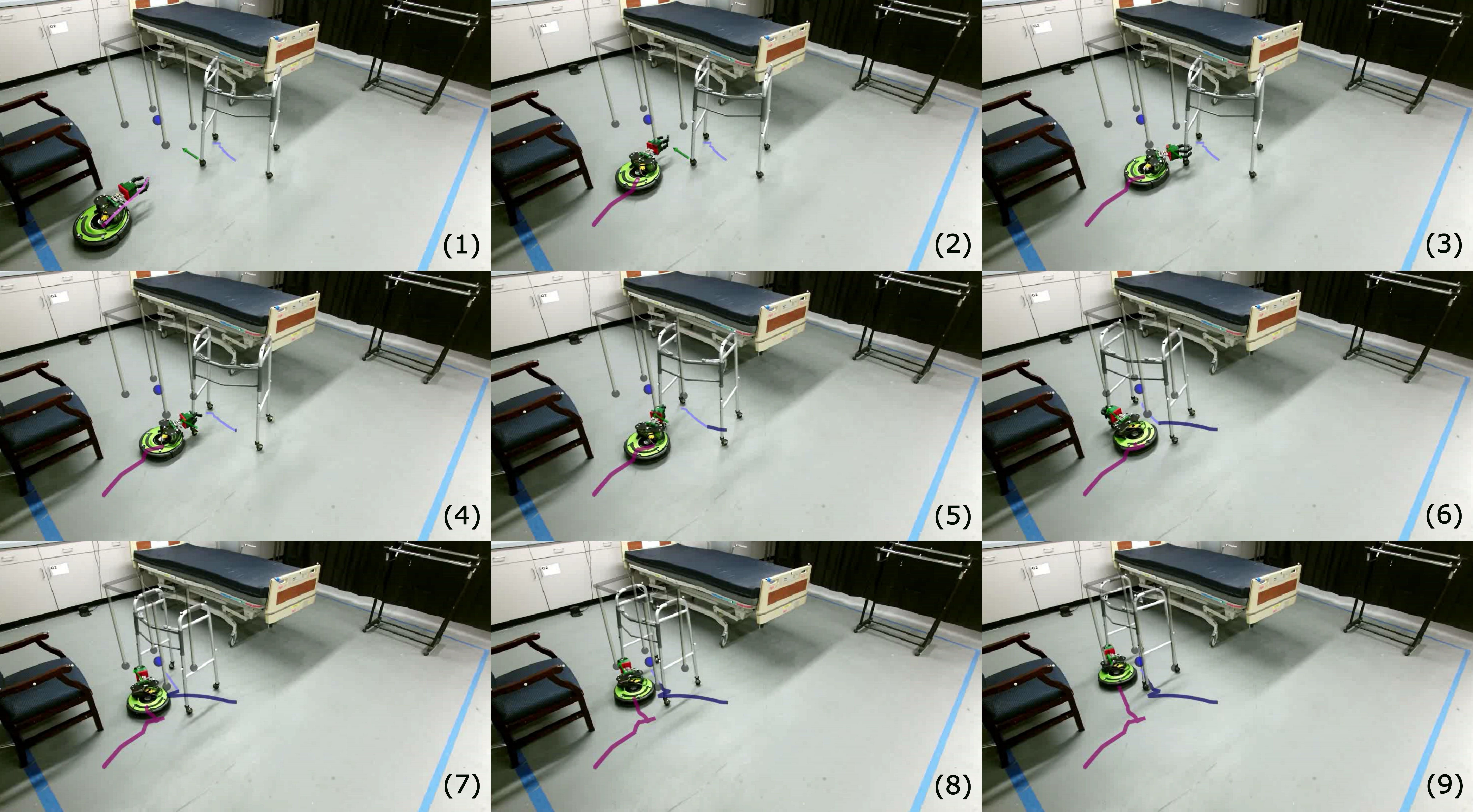}
\caption{Frames of the physical experiment performing the first task. (1) light pink line: robot's planned path, (2) light blue line: object's planned path, (3) dark pink line: robot's actual path, (4) dark blue line: object's actual path. The desire object configuration is also shown as a simulated walker with the blue sphere showing its center.}
\label{frames1}
\vspace{0.3in}
\end{figure*}

\begin{figure*}[t!]
\center
\includegraphics[width = 17.5cm]{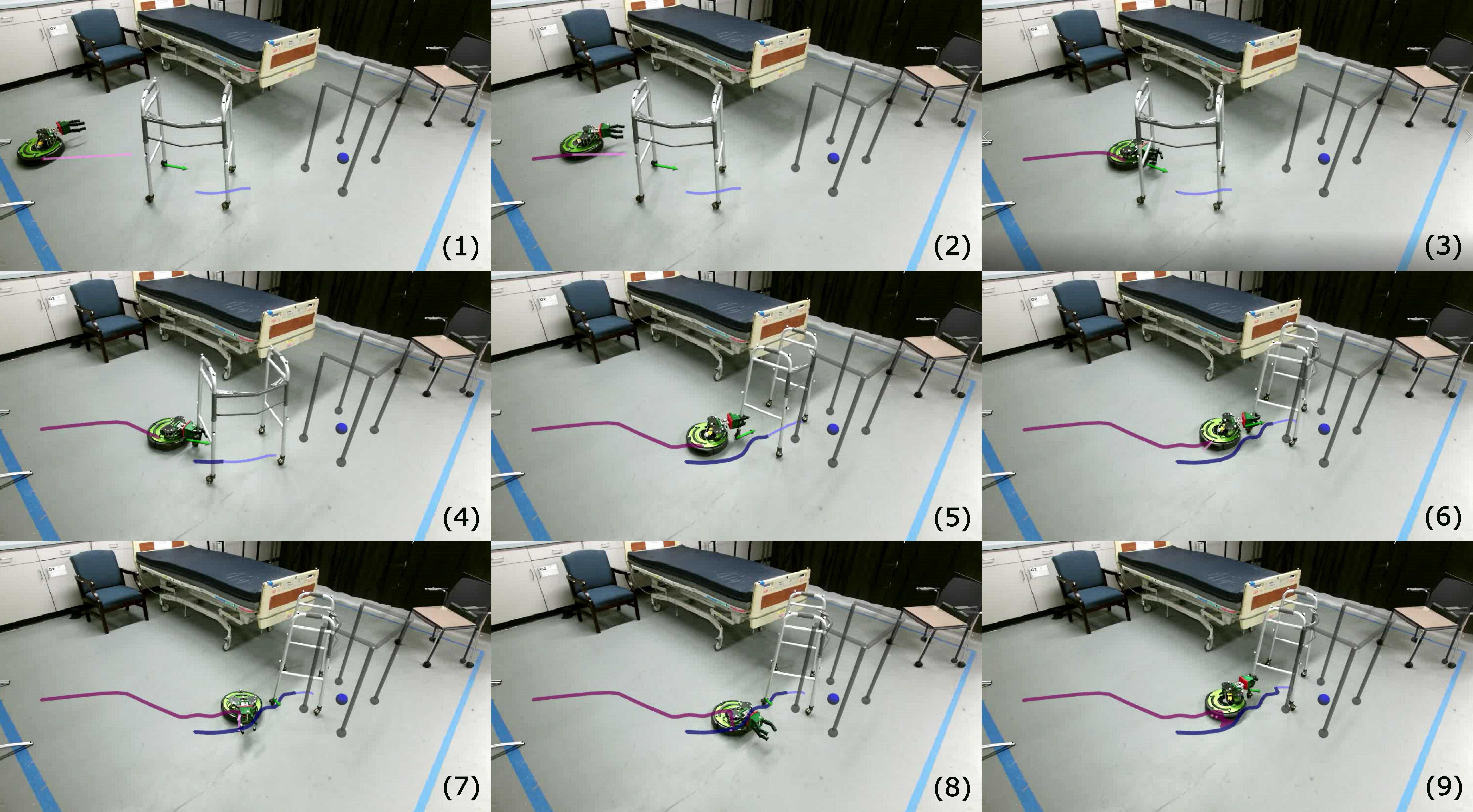}
\caption{Frames of the physical experiment performing the second task. (1) light pink line: robot's planned path, (2) light blue line: object's planned path, (3) dark pink line: robot's actual path, (4) dark blue line: object's actual path. The desire object configuration is also shown as a simulated walker with the blue sphere showing its center. In this task, the robot had to reposition once near the end of the trajectory.}
\label{frames2}
\vspace{0.3in}
\end{figure*}

\begin{figure}[t!]
\center
\hspace{-0.2in}\includegraphics[width = 9cm]{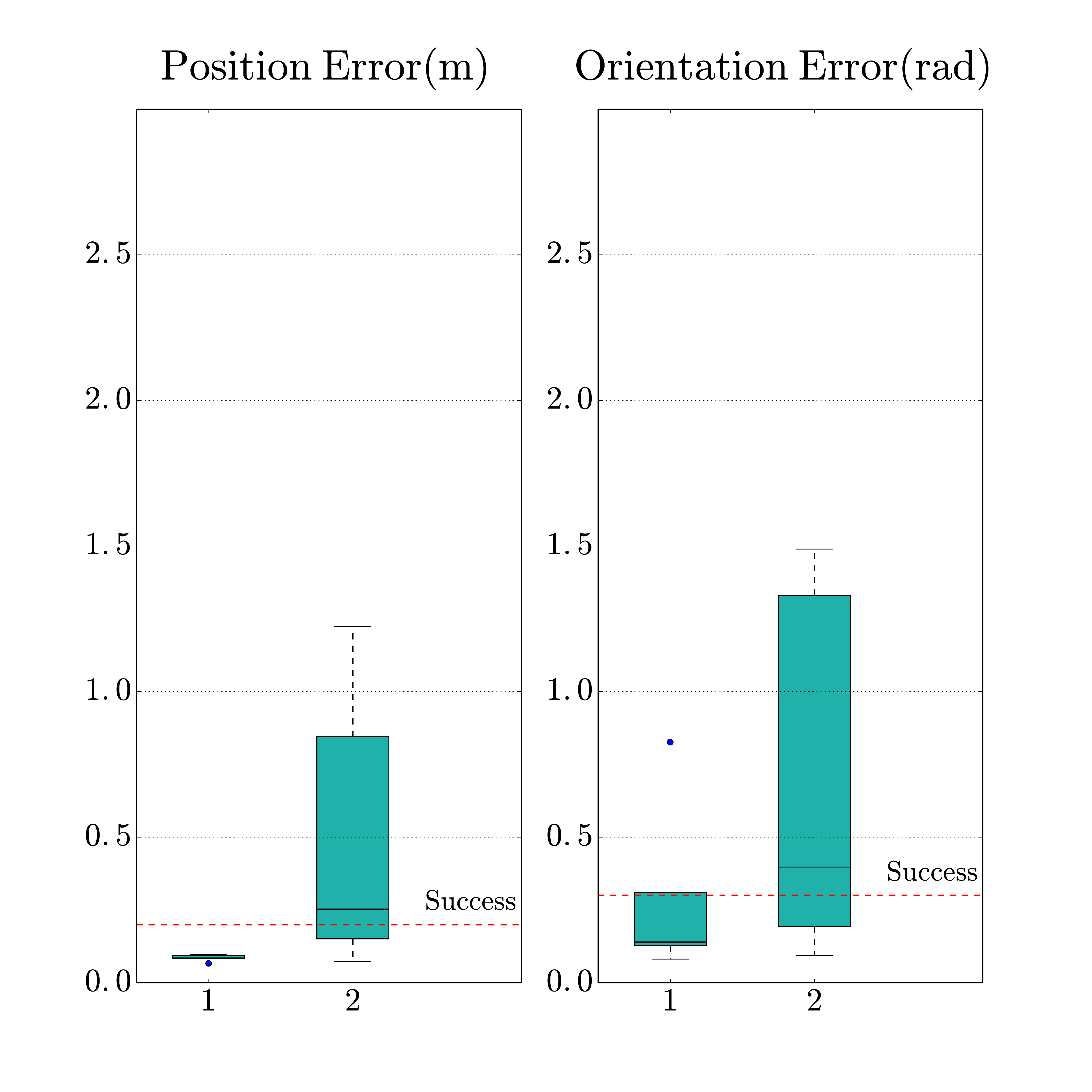}
\vspace{-0.2in}
\caption{Final position and orientation errors for one object in two real-world experiments. The x-axis shows the task number. We assume that the maximum allowed run time in all trials is 3 minutes, and the system stops after that even if it has not reached the goal region yet. The goal region is defined as a distance less than $20$ cm to the goal position with less than $0.3$ rad deviation from the goal orientation, which is shown as a red dashed line.}
\label{real_results}
\end{figure}

\subsection{Manipulation planning}

We ran the manipulation planner for four different simulation setups for all objects (Fig.~\ref{sample_results}). For each simulation trial, the actual dynamic parameter is sampled from the learned distribution. However, in planning, we always use the mean value. We also add noise to the system for simulating the resulting trajectory. For each setup, we compare the results of 50 trials from our approach and LQR using an initial trajectory obtained from optimization. Figure~\ref{sample_results} Shows examples from resulting trajectories for both of these methods for all setups. As we can see, the LQR method is almost never successful due to its inability to compensate for large errors in the system model. Using MPC, the framework lets the system recover from both modeling errors and noise in the system. 

\begin{table}[t!]
\caption{Success rate (\%) in simulation experiments using the proposed method.}
\label{results_sim}
\centering
\begin{tabularx}{\columnwidth}{m{0.25\columnwidth} m{0.15\columnwidth} m{0.15\columnwidth} m{0.15\columnwidth} m{0.15\columnwidth}} 
\hline
Task & 1 & 2 & 3 & 4\\
\hline
Walker & 96 & 66 & 70 & 56\\ 
Blue Chair & 50 & 24 & 50 & 32\\
Gray Chair & 60 & 24 & 38 & 48\\
Rack & 16 & 36 & 0 & 0\\
\hline
\end{tabularx}
\vspace{0.1in}
\end{table}

To see how the proposed method works for all objects, Fig.~\ref{sim_results} represents the final position and orientation errors for all objects through all tasks. We assume that the maximum allowed run time in all trials is 3 minutes, and the system stops after that even if it has not reached the goal region yet. The goal region is defined as a distance less than $10$ cm to the goal position with less than $0.2$ rad deviation from the goal orientation. The success rate using our method is also reported in Table.~\ref{results_sim}. 

As we can see, position errors are generally lower than the orientation errors. This is because our robot can not apply torque directly and has to reach the desired orientation by only applying force. This, along with the limitations in robot motion, makes refining the final orientation very difficult.

In terms of success, the walker has the highest success rate. We believe the reason for this is that we have defined all weights and scenarios using the walker object and used the same weights and scenarios for all other objects without any modification. For example, our last object, which is a rack, is large, and as we can see, two of the scenarios were not suitable for that object at all due to the limited space. In addition, tuning optimization weights can affect the overall performance of the proposed method. With a more systematic weight tuning or an adaptive weight assignment, we can improve the results for other objects. 

The object shape and size, as long as the robot can move it, does not affect the manipulation planning framework, and it can be used for similar objects with various shapes and sizes. However, the physical limitations or the robot used in this study limited the types of objects and their weight and leg radius. The current version of the robot is only designed to grasp leg sizes between 30 mm to 50 mm. In addition, we used objects that are as light weight as possible to avoid large desired forces that are infeasible for the robot to execute.

Moreover, the object's rotational inertia plays an important role in the success of the final orientation. Higher rotational inertia means that the object needs more torque for rotation, which is harder to perform by only applying force to one of the object's legs. We can see this effect in the second object, which is the heaviest object evaluated, a 4-wheeled chair.

Running on a Core i7 2.4GH platform, the computational time for each step is less than a millisecond and the planning rate is 10Hz, which is sufficient for uninterrupted execution in our case. However, this varies based on the feasibility and complexity of the problem.

The robot limitations only allowed for experiments with the walker. To better visualize our proposed method in the real world, we present frames of both tasks in Figs. \ref{frames1} and \ref{frames2}. In these figures, we show planned and actual paths for both the robot and the object, as well as the desired object configuration. The videos of our experiments can be found at \href{https://sites.google.com/view/mobile-manipulation-planning}{https://sites.google.com/view/mobile-manipulation-planning}. 

Results from two tasks, each with 5 trials are reported in Fig.~\ref{real_results} and Table \ref{results_real}. We can see that the second task (40\% success rate) is more difficult than the first one (80\% success rate). We believe this is mainly because of the long distance in the second task, which needs more repositioning actions, resulting in greater error and opportunities to fail. A better re-grasp planning approach would improve performance.

\begin{table}[t!]
\caption{Physical experiment results for two tasks using the walker as target object.}
\label{results_real}
\centering
\begin{tabularx}{\columnwidth}{m{0.5\columnwidth} m{0.2\columnwidth} m{0.3\columnwidth}} 
\hline
Task & 1 & 2  \\
\hline
Success Rate (\%) & 80 & 40\\ 
Average Run time (s) & 125 & 160\\
\hline
\end{tabularx}
\vspace{0.3in}
\end{table}

\section{Conclusion}

We presented an optimization-based framework for mobile manipulation. We focused on the problem of moving large legged objects in which we have to choose between legs to push or pull. We implemented a Bayesian regression method for autonomous learning of approximate dynamic parameters given three different models. We show that a simple ``point mass on a wheel" model is sufficient for our application. However, it is possible to use more complicated models as well. 

We use mixed-integer convex optimization to solve the hybrid control problem comprised of i) choosing which leg to grasp, and ii) continuous applied forces to move the object. Using MPC lets the system recover from modeling errors and find an optimal path to manipulate the object to the desired configuration. We validated our algorithm in simulated problems and real-world experiments. In simulations, we investigate the effect of replanning by comparing our algorithm with LQR.

In the process, we also found that the optimization weights have a significant effect on the performance of planning, and a systematic method to assign those for each object should be found. As future work, we would like to conduct physical experiments with a more powerful mobile robot and use other objects with different wheel configurations. Besides, a better obstacle avoidance approach, which is not as conservative as the Big-M method, would leave more space for the robot to maneuver, probably leading to a higher success rate.

Another interesting possibility for future research is the grasp planning of legged objects considering the manipulation plan. In other words, planning the grasp position so that the robot grasps the leg from the best possible direction to increase the amount of manipulation with the same grasp. This will decrease the need for repositioning.

Finally, this planning approach should be validated with real perception. Moreover, we would like to perform user studies to evaluate the performance of the algorithm in delivering mobility aids to humans.

\bibliographystyle{SageH}
\bibliography{ref.bib}





\end{document}